\documentclass{article}
\usepackage{subcaption}
\usepackage{wrapfig}

\usepackage[final]{neurips_2022}

\usepackage{amsmath,amsfonts,bm}

\def\eqref#1{equation~\ref{#1}}

\def\1{\bm{1}}

\DeclareMathAlphabet{\mathsfit}{\encodingdefault}{\sfdefault}{m}{sl}
\SetMathAlphabet{\mathsfit}{bold}{\encodingdefault}{\sfdefault}{bx}{n}

\newcommand{\E}{\mathbb{E}}

\usepackage[utf8]{inputenc} %
\usepackage[T1]{fontenc}    %
\usepackage{hyperref}       %
\usepackage{url}            %
\usepackage{booktabs}       %
\usepackage{amsfonts}       %
\usepackage{nicefrac}       %
\usepackage{microtype}      %
\usepackage{xcolor}         %
\usepackage{todonotes}
\usepackage[capitalise,noabbrev,nameinlink]{cleveref}
\usepackage{algorithm}
\usepackage{algpseudocode}

\newcommand{\revision}[1]{\textcolor{black}{{#1}}}

\newcommand{\ie}{\emph{i.e.}}
\newcommand{\obs}{\mathbf{o}}
\newcommand{\act}{a}
\newcommand{\rew}{r}
\newcommand{\ret}{R}

\newcommand{\expert}{\text{expert}}

\title{Multi-Game Decision Transformers}

\author{%

Kuang-Huei Lee\thanks{Equal contribution. \texttt{[leekh, ofirnachum, imordatch]@google.com}} \qquad Ofir Nachum$^*$ \qquad Mengjiao Yang \qquad Lisa Lee \AND Daniel Freeman \qquad Winnie Xu \qquad Sergio Guadarrama \qquad Ian Fischer \AND Eric Jang \qquad Henryk Michalewski \qquad Igor Mordatch$^*$ \\
\\
Google Research

}

\begin{document}

\maketitle

\begin{abstract}

A longstanding goal of the field of AI is a method for learning a highly capable, generalist agent from diverse experience.
In the subfields of vision and language, this was largely achieved by scaling up transformer-based models and training them on large, diverse datasets.
Motivated by this progress, we investigate whether the same strategy can be used to produce generalist reinforcement learning agents.
Specifically, we show that a single transformer-based model -- with a single set of weights -- trained purely offline can play a suite of up to 46 Atari games simultaneously at close-to-human performance.
When trained and evaluated appropriately, we find that the same trends observed in language and vision hold, including scaling of performance with model size and rapid adaptation to new games via fine-tuning.
We compare several approaches in this multi-game setting, such as online and offline RL methods and behavioral cloning, and find that our Multi-Game Decision Transformer models offer the best scalability and performance.
\setcounter{footnote}{0}
We release the pre-trained models and code to encourage further research in this direction.\footnote{%
    Additional information, videos and code can be seen at \href{https://sites.google.com/view/multi-game-transformers}{sites.google.com/view/multi-game-transformers}.
}

\end{abstract}

\section{Introduction}
Building large-scale generalist models that solve many tasks by training on massive task-agnostic datasets has emerged as a dominant approach in natural language processing \citep{devlin2018bert,brown2020language}, computer vision \citep{dosovitskiy2020image,arnab2021vivit}, and their intersection \citep{radford2021learning,alayrac2022flamingo}. 
These models can adapt to new tasks (such as translation \citep{raffel2019exploring,xue2021mt5}), make use of unrelated data (such as using high-resource language to improve translations of low-resource languages \citep{dabral2021rudder}), or even incorporate new modalities by projecting images into language space \citep{lu2021pretrained,tsimpoukelli2021multimodal}.
The success of these methods largely derives from a combination of scalable model architectures \citep{vaswani2017attention}, an abundance of unlabeled task-agnostic data, and continuous improvements in high performance computing infrastructure. 
Crucially, scaling laws \citep{kaplan2020scaling,hoffmann2022training} indicate that performance gains due to scale have not yet reached a saturation point.

In this work, we argue that a similar progression is possible in the field of reinforcement learning, and take initial steps toward scalable methods that produce highly capable generalist agents.
In contrast to vision and language domains, reinforcement learning has seen advocacy for the use of smaller models \citep{cuccu2018playing,mania2018simple,bastani2018verifiable} and is usually either used to solve single tasks, or multiple tasks within the same environment.
Importantly, training across multiple environments -- with very different dynamics, rewards, visuals, and agent embodiments -- has been studied less significantly.

Specifically, we investigate whether a single model -- with a single set of parameters -- can be trained to act in multiple environments from large amounts of expert and non-expert experience.
We consider training on a suite of 41 Atari games \citep{bellemare13arcade,gulcehre2020rl} for their diversity, informally asking ``Can models learn something universal from playing many video games?''. 
To train this model, we use only the previously-collected trajectories from \citet{agarwal2020optimistic}, but we evaluate our agent interactively.
We are not striving for mastery or efficiency that game-specific agents can offer, as we believe we are still in early stages of this research agenda. 
Rather, we investigate whether the same trends observed in language and vision hold for large-scale generalist reinforcement learning agents.

\begin{figure}[t]
    \centering
    \includegraphics[width=1.0\linewidth]{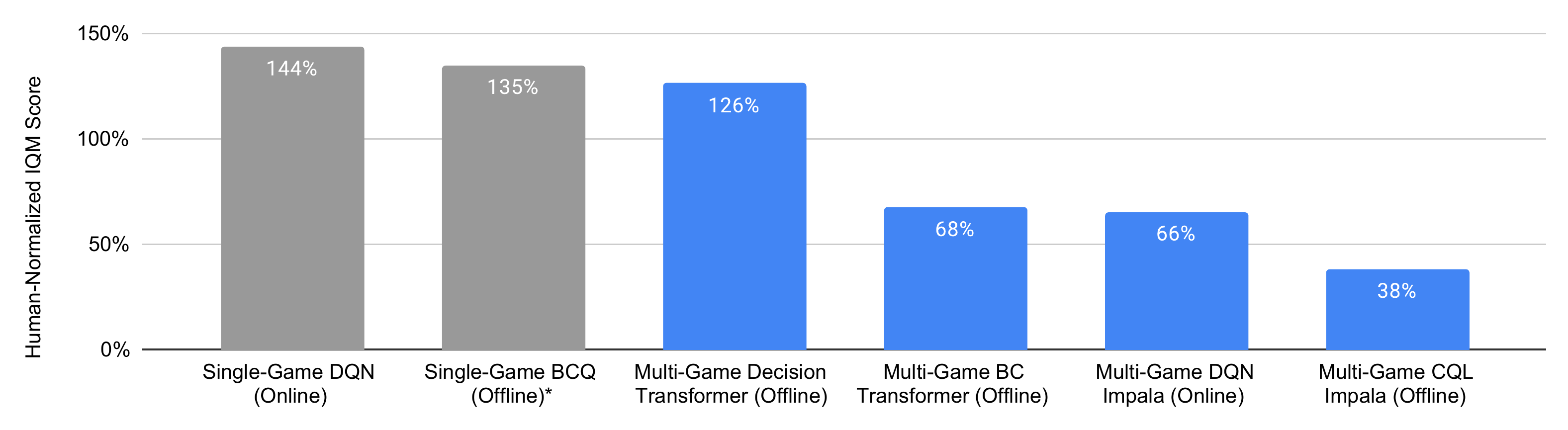}
    \caption{%
        Aggregates of human-normalized scores (Inter-Quartile Mean) across 41 Atari games. 
        Grey bars are single-game specialist models while blue are generalists. 
        Single-game BCQ~\citep{fujimoto2019off} results are from \citet{gulcehre2020rl}.
        Multi-game models are all trained on a dataset~\citep{agarwal2020optimistic} with inter-quartile mean human-normalized score of 101\%, which Multi-Game DT notably exceeds.
    }
    \label{fig:method_scores}
\end{figure}

We find that we can train a single agent that achieves 126\% of human-level performance simultaneously across all games after training on offline expert and non-expert datasets (see \Cref{fig:method_scores}).
Furthermore, we see similar trends that mirror those observed in language and vision: rapid fine-tuning to never-before-seen games with very little data (\Cref{sec:exp_finetune}), a scaling relationship between performance and model size (\Cref{sec:exp_scaling}), and faster training progress for larger models (\cref{sec:model_size}).

Notably, not all existing approaches to multi-environment training work well.
We investigate several approaches, including treating the problem as offline decision transformer-based sequence modeling \citep{chen2021decision, janner2021offline}, online RL \citep{mnih2015human}, offline temporal difference methods \citep{kumar2020conservative}, contrastive representations \citep{oord2018representation}, and behavior cloning \citep{pomerleau1991efficient}. 
We find that decision transformer based models offer the best performance and scaling properties in the multi-environment regime. 
However, to permit training on both expert and non-expert trajectories, we find it is necessary to use a guided generation technique from language modeling to generate expert-level actions, which is an important departure from standard decision transformers. %

Our contributions are threefold:
First, we show that it is possible to train a single high-performing generalist agent to act across multiple environments from offline data alone.
Second, we show that scaling trends observed in language and vision hold.
And third, we compare multiple approaches for achieving this goal, finding that decision transformers combined with guided generation perform the best. 
It is our hope this study can inspire further research in generalist agents. 
To aid this, we make our pre-trained models and code publicly available.

\begin{figure}[ht]
    \centering
    \includegraphics[width=1.0\linewidth]{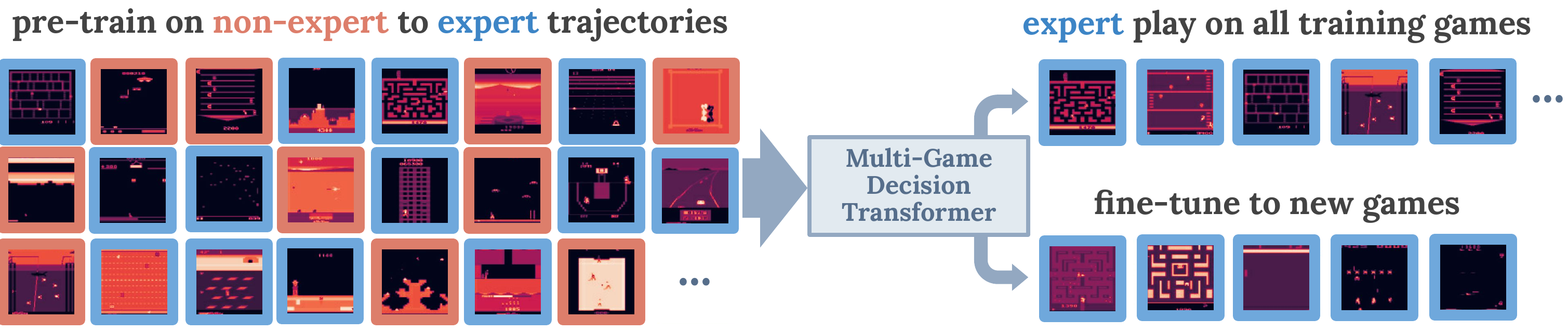}
    \caption{
        An overview of the training and evaluation setup. 
        We observe expert-level game-play in the interactive setting after offline learning from trajectories ranging from beginner to expert.}
    \label{fig:teaser}
\end{figure}

\section{Related Work}
A generalist agent for solving a variety of environments has been a goal for artificial intelligence (AI) researchers since the inception of AI as a field of study~\citep{mccarthy2006proposal}.
This same reason motivated the introduction of the Atari suite (the Arcade Learning Environment, or ALE) as a testbed for learning algorithms~\citep{bellemare2013arcade}; in their own words, the ALE is for ``empirically assessing agents designed for general competency.''
While the celebrated deep $Q$-learning~\citep{mnih2013playing} and actor critic~\citep{mnih2016asynchronous} agents were among the first to use a single algorithm for all games, they nevertheless required separate training and hyperparameters for each game agent.
Later works have demonstrated the ability to learn a single neural network agent on multiple Atari games simultaneously, either online \citep{espeholt2018impala} or via policy distillation~\citep{parisotto2015actor, rusu2015policy}.
The aim of our work is similar -- to learn a single agent for playing multiple Atari games -- with a focus on offline learning.
We demonstrate results with human-level competency on up to 46 games, which is unseen in the literature.

A closely related setting is learning to solve multiple tasks within the same or similar environments.
For example in the robotics field, existing works propose to use language-conditioned tasks~\citep{lynch2020language,ahn2022can,jang2022bc}, while others posit goal-reaching as a way to learn general skills~\citep{mendonca2021discovering}, among other proposals~\citep{kalashnikov2021mt,yu2020meta}.
In this work, we tackle the problem of learning to act in a large collection of environments with distinctively different dynamics, rewards, and agent embodiments.
This complicated but important setting requires a different type of generalization that has been studied significantly less.

A concurrent work~\citep{gato2022deepmind} also aims to train a transformer-based generalist agent based on offline data including for the ALE. 
This work differs from ours in that the offline training data is exclusively near-optimal and it requires prompting by expert trajectories at inference time.
In contrast, we extend decision transformers \citep{chen2021decision} from the Upside-Down RL family~\citep{srivastava2019training,schmidhuber2019reinforcement} to learn from a diverse dataset (expert and non-expert data), predict returns, and pick optimality-conditioned returns.
Furthermore, we provide comparisons against existing behavioral cloning, online and offline RL methods, and contrastive representations~\citep{yang2021representation,oord2018representation}.
Other works that also consider LLM-like sequence modeling for a variety of single control tasks include~\citep{reid2022can,zheng2022online,janner2021offline,furuta2021generalized,ortega2021shaking}.

\section{Method}
\label{sec:method}

We consider a decision-making agent that at every time $t$ receives an observation of the world $\obs^t$, chooses an action $\act^t$, and receives a scalar reward $\rew^t$.
Our goal is to learn a single optimal policy distribution $P^*_\theta(\act^t | \obs^{\leq t}, \act^{<t}, \rew^{<t})$ with parameters $\theta$ that maximizes the agent's total future return $\ret^t = \sum_{k>t} \rew^k$ on all the environments we consider.

\subsection{Reinforcement Learning as Sequence Modeling}
\label{subsec:rl_as_seq}

\begin{figure}[ht]
    \centering
    \includegraphics[width=0.80\linewidth]{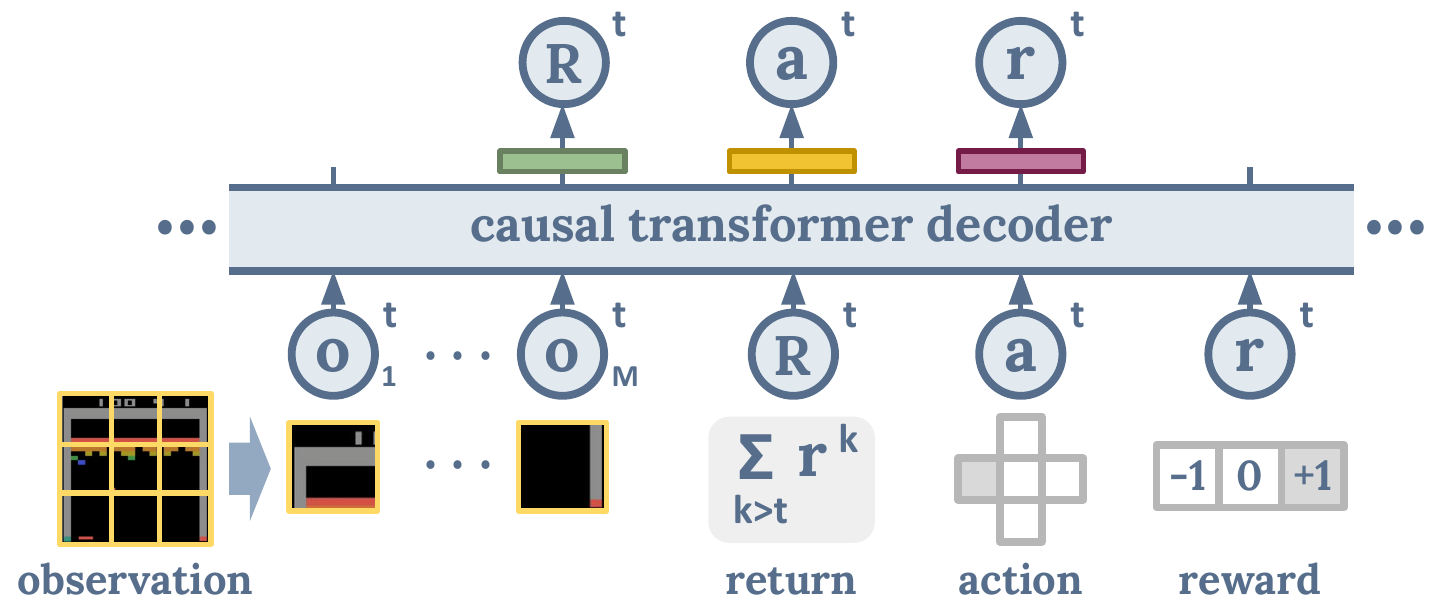}
    \caption{An overview of our decision transformer architecture.}
    \label{fig:model}
\end{figure}

Following \citep{chen2021decision}, we pose the problem of offline reinforcement learning as a sequence modeling problem where we model the probability of the next sequence token $x_i$ conditioned on all tokens prior to it: $P_\theta(x_i|x_{<i})$, similar to contemporary decoder-only sequence models \citep{brown2020language,chowdhery2022palm,rae2021scaling}. 
The sequences we consider have the form:
\begin{equation*}
\label{eq:sequence}
    x \; = \; \langle..., \obs^t_1, ..., \obs^t_M, \hat{\ret}^t, \act^t, \rew^t ,...\rangle
\end{equation*}
where $t$ represents a time-step, $M$ is the number of image patches per observation (which we further discuss in \Cref{subsec:tokenization}), and $\hat{\ret}^t$ is the agent's target return for the rest of the sequence.
Such a sequence order respects the causal structure of the environment decision process.
\Cref{fig:model} presents an overview of our model architecture. 

Returns, actions, and rewards are tokenized (See \Cref{subsec:tokenization} for details), and we train the model to predict the next return, action, and reward discrete token in a sequence via standard cross-entropy loss. 
The sequence we consider is different from \citet{chen2021decision}, which has $\langle..., \hat{\ret}^t, \obs^t, \act^t, ...\rangle$. 
Our design allows predicting the return distribution and sampling from it, instead of relying on a user to manually select an expert-level return at inference time (See \Cref{subsec:expert_action_infer}).

Predicting future value and rewards have been shown to be useful objectives for learning better representations in artificial reinforcement learning agents \citep{lyle2021effect,schrittwieser2020mastering,lee2020predictive} and 
important signals for representation learning in humans \citep{alexander2021representation}.
Thus, while we may not directly use all of the predicted quantities, the task of predicting them encourages structure and representation learning of our environments. 
In this work, we do not attempt to predict future observations due to their non-discrete nature and the additional model capacity that would be required to generate images. However, building image-based forward prediction models of the environment has been shown to be a useful representation objective for RL \citep{hafner2019learning,hafner2019dream,hafner2020mastering}.
We leave it for future investigation.

\subsection{Tokenization}
\label{subsec:tokenization}
To generate returns, actions, and rewards via multinomial distributions similarly to language generation, we convert these quantities to discrete tokens. Actions $\act$ are already discrete quantities in the environments we consider. We convert scalar rewards to ternary quantities $\{-1, 0, +1\}$, and uniformly quantize returns into a discrete range shared by all our environments.\footnote{%
    The training datasets we use (\Cref{subsec:training_dataset}) contains scalar reward values clipped to $[-1, 1]$. For return quantization, we use range $\{-20, ..., 100\}$ with bin size 1 in all our experiments as we find it covers most of the returns we observe in the datasets.
}

Inspired by the simplicity and effectiveness of transformer architectures for processing images \citep{dosovitskiy2020image}, we divide each observation image into a collection of $M$ patches\footnote{%
    We use 6x6 patches, where each patch corresponds to 14x14 pixels, in all our experiments.
} (see \Cref{fig:model}). Each patch is additively combined with a trainable position encoding and linearly projected into the input token embedding space.
We experimented with using image tokenizations coming from a convolutional network, but did not find it to have a significant benefit and omitted it for simplicity.

We chose our tokenization scheme with simplicity in mind, but many other schemes are possible.
While all our environments use a shared action space, varying action spaces when controlling different agent morphologies can still be tokenized using methods of \citep{huang2020one,kurin2020my,gupta2021embodied}.
And while we used uniform quantization to discretize continuous quantities, more sophisticated methods such as VQ-VAE \citep{van2017neural} can be used to learn more effective discretizations. 

\subsection{Training Dataset}
\label{subsec:training_dataset}
To train the model, we use an existing dataset of Atari trajectories (with quantized returns) introduced in \citep{agarwal2020optimistic}. 
The dataset contains trajectories collected from the training progress of a DQN agent \citep{mnih2015human}. 
Following \citep{gulcehre2020rl}, we select 46 games where DQN significantly outperforms a random agent. 
41 games are used for training and 5 games are held out for out-of-distribution generalization experiments. 

We chose 5 held-out games representing different categories including \texttt{Alien} and \texttt{MsPacman} (maze based), \texttt{Pong} (ball tracking), \texttt{SpaceInvaders} (shoot vertically), and \texttt{StarGunner} (shoot horizontally), to ensure out-of-distribution generalization can be evaluated on different types of games.

For each of 41 games, we use data from 2 training runs, each containing roll-outs from 50 policy checkpoints, in turn each containing 1 million environment steps.
This totals 4.1 billion steps. 
Using the tokenization scheme in previous sections, the dataset contains almost 160 billion tokens.

As the dataset contains agent behaviors at all stages of learning, it contains both expert and non-expert behaviors. We do not perform any special filtering, curation, or balancing of the dataset. The motivation to train on such data instead of expert-only behaviors is twofold: Firstly, sub-optimal behaviors are more diverse than optimal behaviors and may still be useful for learning representations of the environment and consequences of poor decisions. 
Secondly, it may be difficult to create a single binary criteria for optimality as it is typically a graded quantity. Thus, instead of assuming only task-relevant expert behaviors, we train our model on all available behaviors, yet generate expert behavior at inference time as described in the next section.

\subsection{Expert Action Inference}
\label{subsec:expert_action_infer}
As described above, our training datasets contain a mix of expert and non-expert behaviors, thus directly generating actions from the model imitating the data is unlikely to consistently produce expert behavior (as we confirm in \Cref{sec:exp_bc}). Instead, we want to control action generation to consistently produce actions of highly-rewarding behavior. This mirrors the problem of discriminator-guided generation in language models, for which a variety of methods have been proposed \citep{krause2020gedi,yang2021fudge,ouyang2022training}. 

We propose an inference-time method inspired by \citep{krause2020gedi} and assume a binary classifier $P(\expert^t|...)$ that identifies whether or not the behavior is expert-level before taking an action at time $t$. Following Bayes' rule, the distribution of expert-level returns at time $t$ is then:
\begin{equation*}
    \revision{P(\expert^t| \ret^t, ...) \propto \exp(\kappa (\ret^t - \ret_{low})/(\ret_{high} - \ret_{low}))}
\end{equation*}
where $\ret_{low}$ is the return lower bound and $\ret_{high}$ is the return upper bound.
Similarly to \citep{shachter1988probabilistic,todorov2006linearly,toussaint2009robot,kappen2012optimal}, we define a binary classifier to be proportional to future return with inverse temperature $\kappa$\footnote{We use $\kappa = 10$ in all our experiments.}:
\begin{equation*}
    P(\expert^t| \ret^t, ...) \equiv \exp(\kappa \ret^t)
\end{equation*}

\begin{figure}[h]
    \centering
    \includegraphics[width=0.9\linewidth]{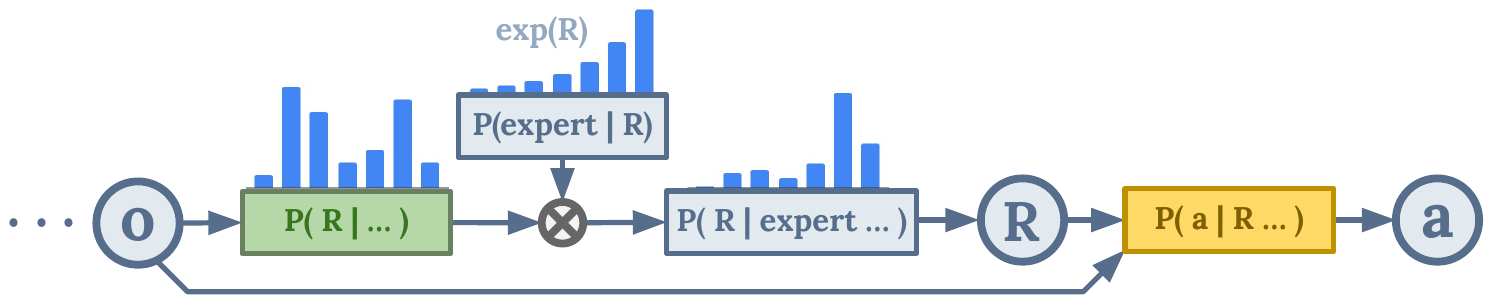}
    \caption{
        An illustration of our expert-level return and action sampling procedure. $P_\theta(\ret |...)$ and $P_\theta(\act|\ret...)$ are the distributions learned by the sequence model.
    }
    \label{fig:inference}
\end{figure}

This results in a simple auto-regressive procedure where we first sample high-but-plausible target returns $R^t$ according to log-probability $\log P_\theta(\ret^t|...) + \kappa (\ret^t - \ret_{low})/(\ret_{high} - \ret_{low})$, and then sample actions according to $P_\theta (\act^t | \ret^t, ...)$. See \Cref{fig:inference} for an illustration of this procedure and \cref{subsec:appendix_eval} for implementation details. It can be seen as a variation of return-conditioned policies \citep{kumar2019reward,srivastava2019training,chen2021decision} that automatically generates expert-level (but likely) returns at every timestep, instead of manually fixing them for the duration of the episode.

Importantly, this formulation only affects the inference procedure of the model -- training is entirely unaffected and can rely on standard next-token prediction frameworks and infrastructure. While we chose this formulation for its simplicity, controllable generation is an active area of study and we expect other more effective methods to be introduced in the future. As such, our contribution is to point out a connection between problems of controllable generation in language modeling and optimality conditioning in control.

\section{Experiments}

We formulate our experiments to answer a number of questions that are addressed in following sections:
\begin{itemize}
\item How do different {\bf online and offline} methods perform in the multi-game regime?
\item How do different methods {\bf scale with model size}?
\item How effective are different methods at {\bf transfer to novel games}?
\item Does {\bf multi-game decision transformer} improve upon training data?
\item Does {\bf expert action inference} (\Cref{subsec:expert_action_infer}) improve upon behavioral cloning?
\item Does training on {\bf expert and non-expert data} bring benefits over expert-only training?
\end{itemize}
We also consider whether there are benefits to specifically using the {\bf transformer architecture} in \cref{sec:udrl}, and qualitatively explore the attention behavior of these models in \cref{sec:attention}.

\subsection{Setup}
\label{sec:exp_setup}

\paragraph{Model Variants and Scaling.}
We base our decision transformer (DT) configuration on GPT-2 \citep{brown2020language} as summarized in \cref{subsec:architecture}. %
We report results for DT-200M (a Multi-Game DT with 200M parameters) if not specified otherwise. 
Other smaller variants are DT-40M and DT-10M.
We set sequence length to 4 game frames for all experiments, which results in sequences of 156 tokens.

\paragraph{Training and Fine-tuning.}
We train all Multi-Game DT models on TPUv4 hardware and the Jaxline (\citet{deepmind2020jax}) framework for 10M steps using the LAMB optimizer \citep{you2019large} with a $3\cdot10^{-4}$ learning rate, 4000 steps linear warm-up, no weight decay, gradient clip 1.0, $\beta_1=0.9$ and $\beta_2=0.999$, and batch size 2048.
For fine-tuning on novel games, we train for 100k steps with a $10^{-4}$ learning rate, $10^{-2}$ weight decay and batch size of 256 instead. Both regimes used image augmentations as described in \cref{subsec:image_aug}.

\paragraph{Metrics.}
We measure performance on individual Atari games by human normalized scores (HNS) \citep{mnih2015human}, i.e. $(\textrm{score} - \textrm{score}_\textrm{random})/(\textrm{score}_\textrm{human} - \textrm{score}_\textrm{random})$, or DQN-normalized scores, i.e. normalizing by the best DQN scores seen in the training dataset instead of using human scores.
To create an aggregate comparison metric across all games, we use inter-quartile mean (IQM) of human-normalized scores across all games, following evaluation best practices proposed in \citep{agarwal2021deep}. 
Due to the prohibitively long training times, we only evaluated one training seed. We additionally report median aggregate metric in \cref{sec:median_scores}.

\subsection{Baseline Methods}

\paragraph{BC}
Our Decision Transformer (\Cref{subsec:rl_as_seq}) can be reduced to a transformer-based Behavioral Cloning (BC)~\citep{pomerleau1991efficient} agent by removing the target return condition and return token prediction.
Similar to what we do for Decision Transformer, we also learn BC models at different scales (10M, 40M, 200M parameters) while keeping other configurations unchanged.

\paragraph{C51 DQN}
As a point of comparison for online performance, we use the C51 algorithm~\citep{bellemare2017distributional} which is a variant of deep $Q$-learning (DQN) but with a categorical loss for minimizing the temporal difference (TD) errors. 
Following improvements suggested in~\citet{hessel2018rainbow} as well as our own empirical observations, we use multi-step learning with $n=4$.
For the single-game experiments, we use the standard convolutional neural network (CNN) used in the implementation of C51~\citep{castro2018dopamine}.
For the multi-game experiments, we modify the C51 implementation based on a hyperparameter search to use an Impala neural network architecture~\citep{espeholt2018impala} with three blocks using $64$, $128$, and $128$ channels respectively with a batch size of $128$ and update period of $256$.

\paragraph{CQL}
For an offline TD-based learning algorithm we use conservative $Q$-learning (CQL)~\citep{kumar2020conservative}. 
Namely, we augment the categorical loss of C51 with a behavioral cloning loss minimizing $-\log \pi_Q(a | s)$, where $(s,a)$ is a state-action pair sampled from the offline dataset and $\pi_Q(\cdot | s) = \mathrm{softmax}(Q(s,\cdot))$. 
Following the recommendations in~\citet{kumar2020conservative} we weight the contribution of the BC loss by $1$ when using 100\% of the offline data (multi-game training) and $4$ when using 1\% (single-game finetuning). 
For scaling experiments, we vary the number of blocks and channels in each block of the Impala: the number of blocks and channels is one of $(5~\mathrm{blocks}, 128~\mathrm{channels})\approx 5\mathrm{M~params}$, $(10~\mathrm{blocks}, 256~\mathrm{channels})\approx 30\mathrm{M~params}$,
$(5~\mathrm{blocks}, 512~\mathrm{channels})\approx 60\mathrm{M~params}$,
$(10~\mathrm{blocks}, 512~\mathrm{channels})\approx 120\mathrm{M~params}$.

\paragraph{CPC, BERT, and ACL}
For rapid adaptation to new games via fine-tuning, we consider representation learning baselines including contrastive predictive coding (CPC)~\citep{oord2018representation}, BERT pretraining~\citep{devlin2018bert}, and attentive contrastive learning (ACL)~\citep{yang2021representation}.
All state representation networks are implemented as additional multi-layer perceptrons (MLPs) or transformer layers on top of the Impala CNN used in C51 and CQL baselines. 
CPC uses two additional MLP layers with $512$ units each interleaved with \texttt{ReLU} activation to represent $\phi(s)$, which is optimized by maximizing $\phi(s)^\top W \phi(s')$ of true transitions $(s,s')$ and minimizing $\phi(s)^\top W \phi(\tilde s)$ where $\tilde s$ is a state randomly sampled from the batch (including states from other games). 
For BERT pretraining, we use $2$ self-attention layers with $4$ attention heads of $256$ units each and feed-forward dimension $512$, and train $\phi(s)$ using BERT's masked self-prediction loss on a trajectory of sequence length $16$. 
ACL shares the same model parametrization as BERT, with the inclusion of action prediction in the pretraining objective.

\subsection{How do different online and offline methods perform in the multi-game regime?}

We compare different online and offline algorithms in the multi-game regime and their single-game counterparts in ~\cref{fig:method_scores}. 
We find that single-game specialists are still most performant. 
Among multi-game generalist models, our Multi-Game Decision Transformer model comes closest to specialist performance. 
Multi-game online RL with non-transformer models comes second, while we struggled to get good performance with offline non-transformer models. We note that our multi-game online C51 DQN median score of 68\% (see \cref{sec:median_scores}) which compares similarly to multi-game median Impala score of 70\%, which we calculated from results reported by \citep{espeholt2018impala} for our suite of games. 

We believe the apparent advantage of offline DT compared to online multi-game methods like C51 may be explained in part through classical differences between online and offline settings in RL \citep{levine2020offline}.
Online methods must balance exploration with the ability to learn and generalize from experience, which could be challenging in the multi-game setting, whereas offline DT only needs to learn to distill and generalize from the fixed multi-game experience given to it (collected by specialist DQN agents \citep{agarwal2020optimistic}).
Beyond the difference between online and offline, one could also argue that C51 suffers from more training instability than DT due to the use of a temporal difference (TD) loss, which we discuss in the next paragraph.

\subsection{How do different methods scale with model size?}
\label{sec:exp_scaling}

In large language and vision models, lowest-achievable training loss typically decreases predictably with increasing model size.  
\citet{kaplan2020scaling} demonstrated an empirical scaling relationship between the capacity of a language model (NLP terminology for a next-token autoregressive generative model) and its performance (negative log likelihood on held-out data).  
These trends were verified over many orders of magnitude of model size, ranging from few-million parameter models to hundreds of billion parameter models.

\begin{figure}[ht]
    \centering
     \begin{subfigure}{0.56\textwidth}
         \centering
         \includegraphics[width=1\linewidth]{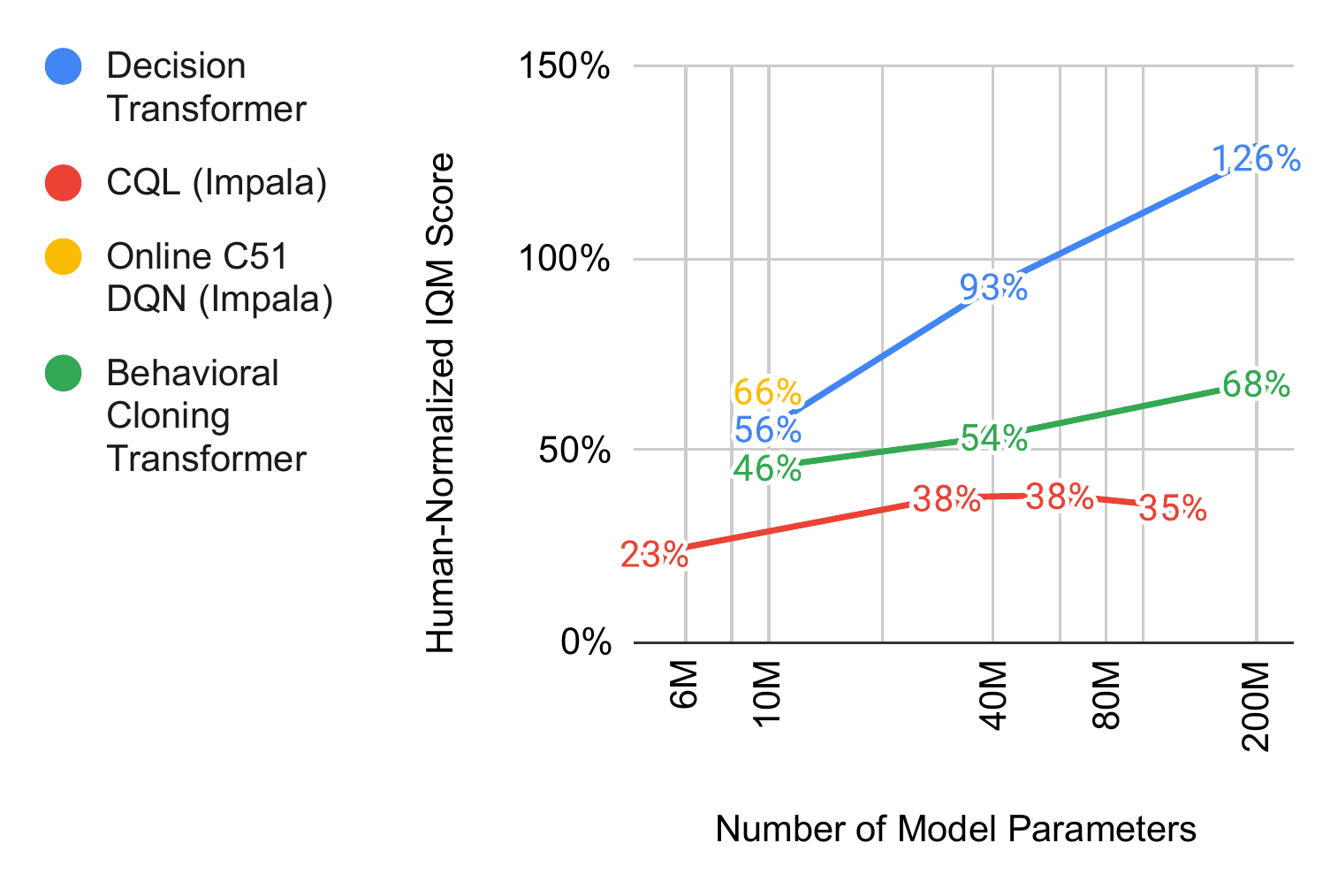}
         \caption{Scaling of IQM scores for all training games with different model sizes and architectures.}
         \label{fig:scaling}         
     \end{subfigure}
     \hfill
     \begin{subfigure}{0.405\textwidth}
         \centering
         \includegraphics[width=1\linewidth]{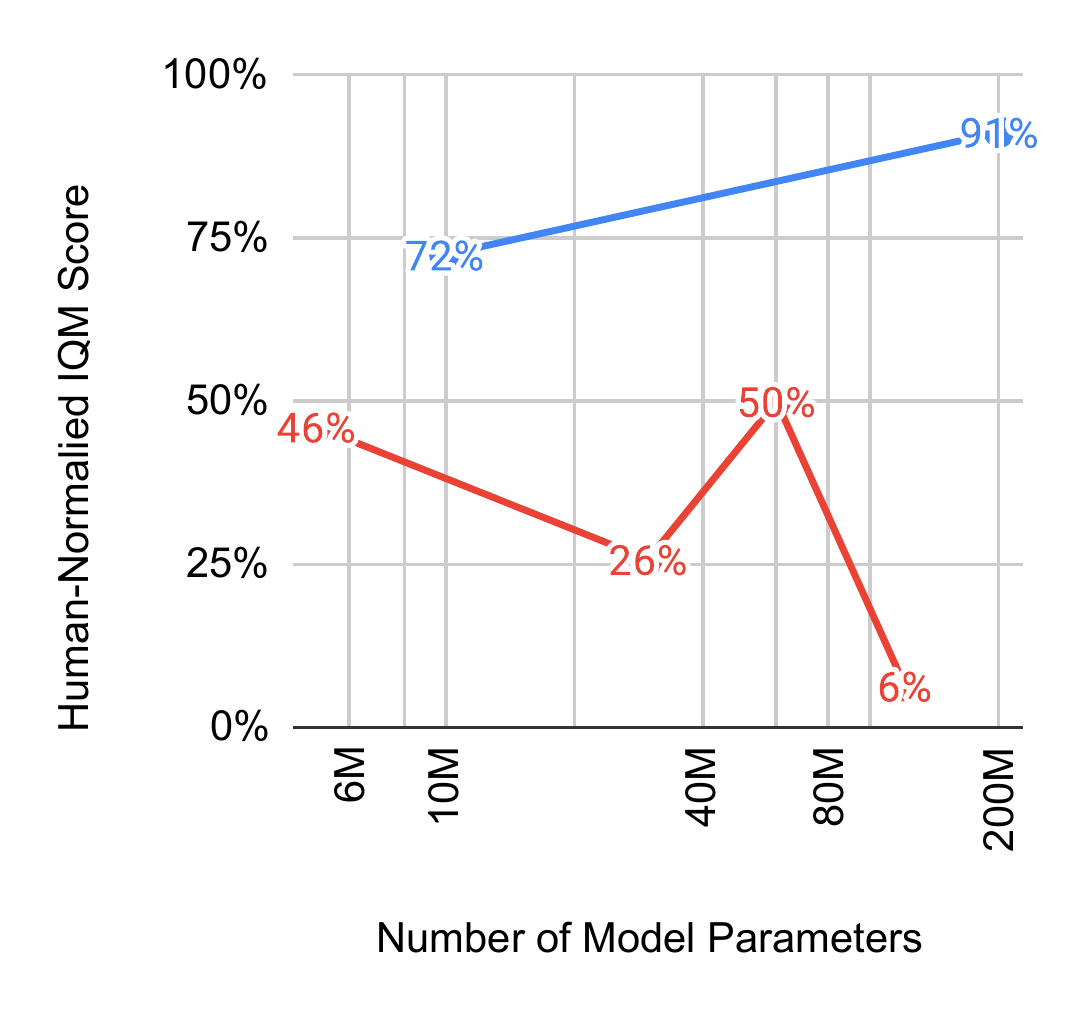}
         \caption{Scaling of IQM scores for all novel games after fine-tuning DT and CQL.} %
         \label{fig:scaling_finetune}             
     \end{subfigure}
     \caption{How model performance scales with model size, on training set games and novel games. 
     (Impala) indicates using the Impala CNN architecture.
     }
\end{figure}

We investigate whether similar trends hold for \emph{interactive} in-game performance -- not just training loss -- and show a similar performance scaling trend in \Cref{fig:scaling}.
Multi-Game Decision Transformer performance reliably increases over more than an order of magnitude of parameter scaling, whereas the other methods either saturate, or have much slower performance growth.

In contrast, in \Cref{fig:scaling,fig:scaling_finetune}, we find that CQL does not improve with increased model size, and actually shows a sharp drop in the performance of larger models on the fine-tuning tasks.
Temporal Difference (TD) methods suffer greater instability with larger model size in the multi-game setting, leading to this ``inverse'' scaling.
Indeed, our attempts at other objectives closer to pure TD (C51, DQN, DDQN) led to even worse results (which we do not report).
We note that similar conclusions about instability with respect to network size have been made by other work~\citep{furuta2021co}.

We also find that larger models train faster, in the sense of reaching higher in-game performance after observing the same number of tokens. 
We discuss these results in \cref{sec:model_size}.

\subsection{How effective are different methods at transfer to novel games?}
\label{sec:exp_finetune}

Pretraining for rapid adaptation to new games has not been explored widely on Atari games despite being a natural and well-motivated task due to its relevance to how humans transfer knowledge to new games. 
\citet{nachum2021provable} employed pretraining on large offline data and fine-tunining on small expert data for Atari and compared to a set of state representation learning objectives based on bisimulation~\citep{gelada2019deepmdp,zhang2020learning}, but their pretraining and fine-tuning use the same game. 
We are instead interested in the \emph{transfer} ability of pretrained agents to new games. 

We hence devise our own evaluation setup by pretraining DT, CQL, CPC, BERT, and ACL on the full datasets of the 41 training games with 100M steps each, and fine-tuning one model per held-out game using 1\% (1M steps) from each game. 
The 1\% fine-tuning data is uniformly sampled from the 50M step dataset without quality filtering. 
DT and CQL use the same objective for pretraining and fine-tuning, whereas CPC, BERT, and ACL each use their own pretraining objective and are fine-tuned using the BC objective. 
All methods are fine-tuned for 100,000 steps, which is much shorter than training any agent from scratch. 
We additionally include training CQL from scratch on the 1\% held-out data to highlight the benefit of rapid fine-tuning.

\begin{figure}[ht]
    \centering
    \includegraphics[width=1.0\linewidth]{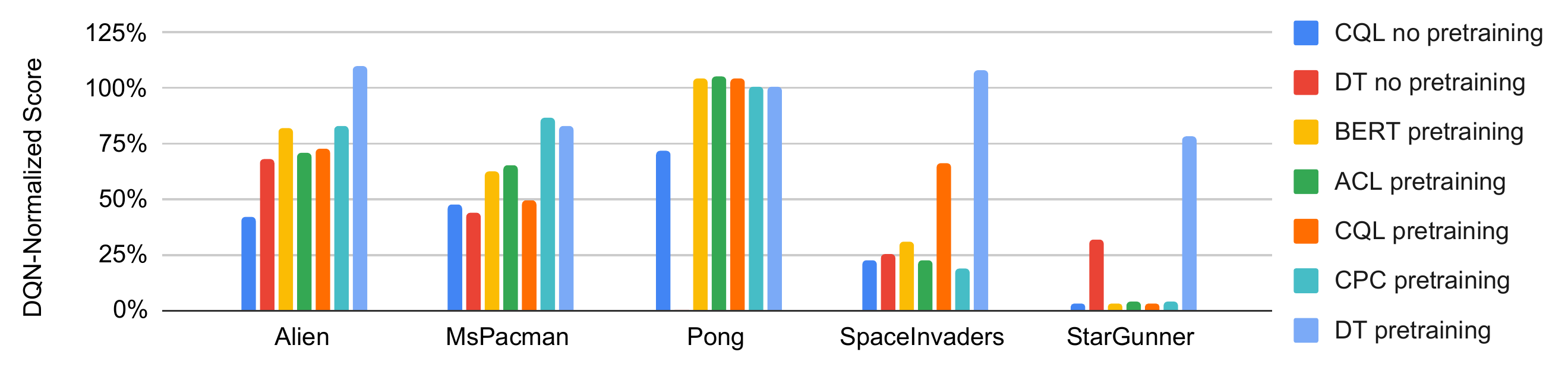}
    \caption{
        Fine-tuning performance on 1\% of 5 held-out games' data after pretraining on other 41 games using DT, CQL, CPC, BERT, and ACL. 
        All pretraining methods outperform training CQL from scratch on the 1\% held-out data, highlighting the transfer benefit of pretraining on other games. 
        DT performs the best among all methods considered.
    }
    \label{fig:finetune}
\end{figure}

Fine-tuning performance on the held-out games is shown in \Cref{fig:finetune}. 
Pretraining with the DT objective performs the best across all games. 
All methods with pretraining outperform training CQL from scratch, which verifies our hypothesis that pretraining on other games should indeed help with rapid learning of a new game. 
CPC and BERT underperform DT, suggesting that learning state representations alone is not sufficient for desirable transfer performance. 
While ACL adds an action prediction auxiliary loss to BERT, it showed little effect, suggesting that modeling the actions in the right way on the offline data is important for good transfer performance. 
Furthermore, we find that fine-tuning performance improves as the DT model becomes larger, while CQL fine-tuning performance is inconsistent with model size (see \cref{fig:scaling_finetune}).

\subsection{Does multi-game decision transformer improve upon training data?}
\begin{figure}[ht]
    \centering
    \includegraphics[width=1.0\linewidth]{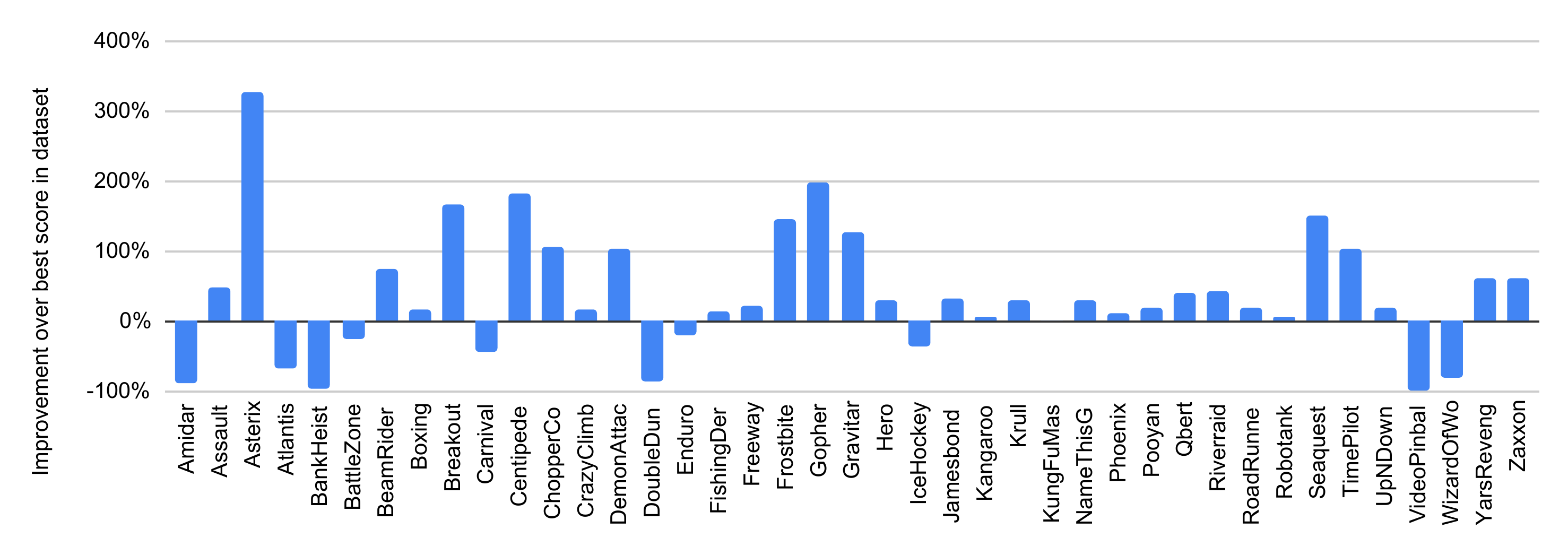}
    \caption{%
        Percent of improvement of top 3 decision transformer rollouts over the best score in the training dataset. 
        0\% indicates no improvement. Top-3 metric (instead of mean) is used to more fairly compare to the best -- rather than expert average -- demonstration score.
    }
    \label{fig:max_improvement}
\end{figure}

We want to evaluate whether decision transformer with expert action inference is capable of acting better than the best demonstrations seen during training. To do this, we look at the top 3 performing decision transformer model rollouts. We use top 3 rollouts instead of the mean across all rollouts to more fairly compare to the \emph{best} demonstration, rather than an average expert demonstration. We show percentage improvement over best demonstration score for individual games in \Cref{fig:max_improvement}. We see significant improvement over the training data in a number of games.

\subsection{Does optimal action inference improve upon behavior cloning?}
\label{sec:exp_bc}

\begin{figure}[ht]
    \centering
    \includegraphics[width=1.0\linewidth]{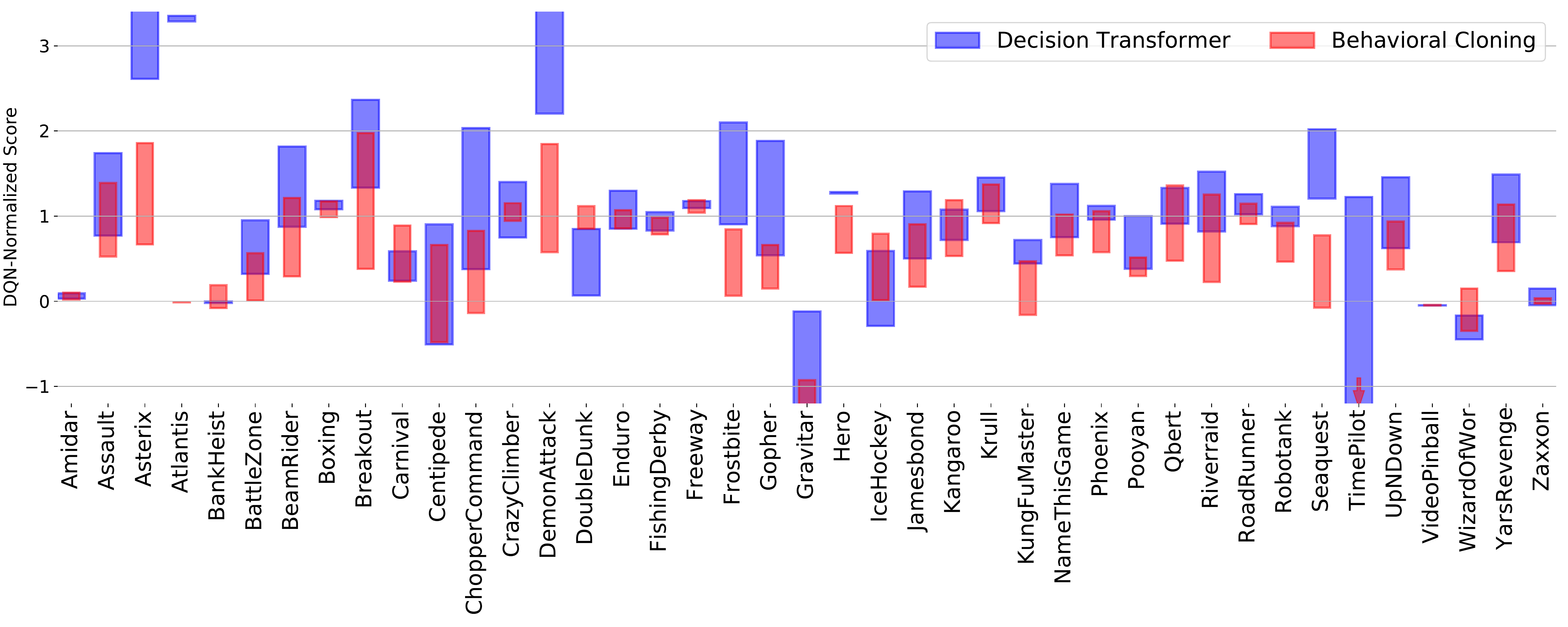}
    \caption{%
        Comparison of per-game scores for decision transformer to behavioral cloning. 
        Bars indicate $\pm$ standard deviation around the mean across 16 trials. 
        We show DQN-normalized scores in this figure for better presentations.
    }
    \label{fig:dt_vs_bc}
\end{figure}

In \Cref{fig:method_scores} we see that IQM performance across all games is indeed significantly improved by generating optimality-conditioned actions. 
\Cref{fig:dt_vs_bc} shows the mean and standard deviation of scores across all games. 
While behavior cloning may sometimes produce highly-rewarding episodes, it is less likely to do so.
We find decision transformer outperforms behavioral cloning in 31 out of 41 games.

\subsection{Does training on expert and non-expert data bring benefits over expert-only training?}
\label{sec:exp_filtering}
\begin{figure}[ht]
     \centering
     \includegraphics[width=1.0\linewidth]{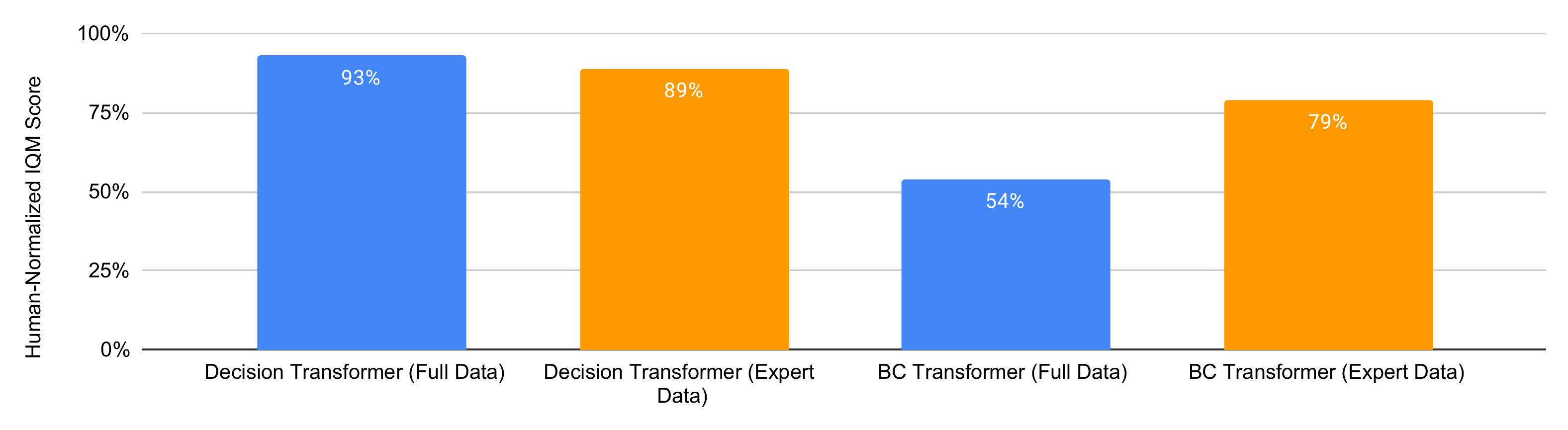}
     \caption{Comparison of 40M transformer models trained on full data and only expert data.}
     \label{fig:expert_iqm_comparison}
\end{figure}

We believe that, comparing to learning from expert demonstrations, learning from large, diverse datasets that include some expert data but primarily non-expert data help learning and improve performance.
To verify this hypothesis, we filter our training data \cite{agarwal2020optimistic} from each game by episodic returns and only preserve top 10\% trajectories to produce an expert dataset (see \cref{sec:filtering_appendix} for details).
We use this expert dataset to train our multi-game decision transformer (DT-40M) and the transformer-based behavioral cloning model (BC-40M).
\Cref{fig:expert_iqm_comparison} compares these models trained on expert data and our DT-40M trained on all data.

We observe that (1) Training only on expert data improves behavioral cloning; (2) Training on full data, including expert and non-expert data, improves Decision Transformer; (3) Decision Transformer with full data outperforms behavioral cloning trained on expert data.

\section{Conclusion}
\label{sec:conclusion}
In the quest to develop highly capable and generalist agents, we have made important and measurable progress.
Namely, our results exhibit a clear benefit of using large transformer-based models in multi-game domains, and the general trends in these results -- performance improvements with larger models and the ability to rapidly fine-tune to new tasks -- mirror the successes observed for large-scale vision and language models.
Our results also highlight difficulties of online RL algorithms in handling the complexity of multi-game training on Atari. 
It is interesting to note that our best results are achieved by decision transformers, which essentially learn via supervised learning on sequence data, compared to alternative approaches such as temporal difference learning (more typical in reinforcement learning), policy gradients, and contrastive representation learning. 
This begs the question of whether online learning algorithms can be modified to be as ``data-absorbent'' as DT-like methods.
While even our best generalist agents at times fall short of performance achieved by agents trained on a single task, this is broadly consistent with related works that have trained single models on many tasks \cite{kaiser2017one, gato2022deepmind}. 
Still, our best generalist agents are already capable of outperforming the data they are trained on.
We believe the trends suggest clear paths for future work -- that, with larger models and larger suites of tasks, performance is likely to scale up commensurately. 

\paragraph{Limitations.} 
We acknowledge reasons for caution in over-generalizing our conclusions. 
Our results are based largely on performance in the Atari suite, where action and observation spaces are aligned across different games. It is unclear whether offline RL datasets such as Atari are of sufficient scale and diversity that we would see similar performance scaling as observed in NLP and vision benchmarks.
Whether we can observe other forms of generalization, such as zero-shot adaptation,
as well as whether our conclusions hold for other settings, remains unclear. 

\paragraph{Societal Impacts.}
In the current setting, we do not foresee significant societal impact as the models are limited to playing simple video games.
We emphasize that our current agents are not intended to interact with humans or be used outside of self-contained game-playing domains. 
One should exercise increased caution if extending our algorithms and methods to such situations in order to ensure any safety and ethical concerns are appropriately addressed. 
At the same time, the capability of decision making based on reward feedback -- rather than purely imitation of the data -- has the potential to be easier to align with human values and goals.

\section*{Acknowledgements}
We would like to thank Oscar Ramirez, Roopali Vij, Sabela Ramos, Rishabh Agarwal, Shixiang (Shane) Gu, Aleksandra Faust, Noah Fiedel, Chelsea Finn, Sergey Levine, John Canny, Kimin Lee, Hao Liu, Ed Chi, and Luke Metz for their valuable contributions and support for this work.

{\small
\bibliographystyle{plainnat}
\bibliography{main}

\begin{thebibliography}{84}
\providecommand{\natexlab}[1]{#1}
\providecommand{\url}[1]{\texttt{#1}}
\expandafter\ifx\csname urlstyle\endcsname\relax
  \providecommand{\doi}[1]{doi: #1}\else
  \providecommand{\doi}{doi: \begingroup \urlstyle{rm}\Url}\fi

\bibitem[Agarwal et~al.(2020)Agarwal, Schuurmans, and
  Norouzi]{agarwal2020optimistic}
Rishabh Agarwal, Dale Schuurmans, and Mohammad Norouzi.
\newblock An optimistic perspective on offline reinforcement learning.
\newblock In \emph{International Conference on Machine Learning}, pages
  104--114. PMLR, 2020.

\bibitem[Agarwal et~al.(2021)Agarwal, Schwarzer, Castro, Courville, and
  Bellemare]{agarwal2021deep}
Rishabh Agarwal, Max Schwarzer, Pablo~Samuel Castro, Aaron~C Courville, and
  Marc Bellemare.
\newblock Deep reinforcement learning at the edge of the statistical precipice.
\newblock \emph{Advances in Neural Information Processing Systems}, 34, 2021.

\bibitem[Ahn et~al.(2022)Ahn, Brohan, Brown, Chebotar, Cortes, David, Finn,
  Gopalakrishnan, Hausman, Herzog, et~al.]{ahn2022can}
Michael Ahn, Anthony Brohan, Noah Brown, Yevgen Chebotar, Omar Cortes, Byron
  David, Chelsea Finn, Keerthana Gopalakrishnan, Karol Hausman, Alex Herzog,
  et~al.
\newblock Do as i can, not as i say: Grounding language in robotic affordances.
\newblock \emph{arXiv preprint arXiv:2204.01691}, 2022.

\bibitem[Alayrac et~al.(2022)Alayrac, Donahue, Luc, Miech, Barr, Hasson, Lenc,
  Mensch, Millican, Reynolds, et~al.]{alayrac2022flamingo}
Jean-Baptiste Alayrac, Jeff Donahue, Pauline Luc, Antoine Miech, Iain Barr,
  Yana Hasson, Karel Lenc, Arthur Mensch, Katie Millican, Malcolm Reynolds,
  et~al.
\newblock Flamingo: a visual language model for few-shot learning.
\newblock \emph{arXiv preprint arXiv:2204.14198}, 2022.

\bibitem[Alexander and Gershman(2021)]{alexander2021representation}
William~H Alexander and Samuel~J Gershman.
\newblock Representation learning with reward prediction errors.
\newblock \emph{arXiv preprint arXiv:2108.12402}, 2021.

\bibitem[Arnab et~al.(2021)Arnab, Dehghani, Heigold, Sun, Lu{\v{c}}i{\'c}, and
  Schmid]{arnab2021vivit}
Anurag Arnab, Mostafa Dehghani, Georg Heigold, Chen Sun, Mario Lu{\v{c}}i{\'c},
  and Cordelia Schmid.
\newblock Vivit: A video vision transformer.
\newblock In \emph{Proceedings of the IEEE/CVF International Conference on
  Computer Vision}, pages 6836--6846, 2021.

\bibitem[Babuschkin et~al.(2020)Babuschkin, Baumli, Bell, Bhupatiraju, Bruce,
  Buchlovsky, Budden, Cai, Clark, Danihelka, Fantacci, Godwin, Jones, Hennigan,
  Hessel, Kapturowski, Keck, Kemaev, King, Martens, Mikulik, Norman, Quan,
  Papamakarios, Ring, Ruiz, Sanchez, Schneider, Sezener, Spencer, Srinivasan,
  Stokowiec, and Viola]{deepmind2020jax}
Igor Babuschkin, Kate Baumli, Alison Bell, Surya Bhupatiraju, Jake Bruce, Peter
  Buchlovsky, David Budden, Trevor Cai, Aidan Clark, Ivo Danihelka, Claudio
  Fantacci, Jonathan Godwin, Chris Jones, Tom Hennigan, Matteo Hessel, Steven
  Kapturowski, Thomas Keck, Iurii Kemaev, Michael King, Lena Martens, Vladimir
  Mikulik, Tamara Norman, John Quan, George Papamakarios, Roman Ring, Francisco
  Ruiz, Alvaro Sanchez, Rosalia Schneider, Eren Sezener, Stephen Spencer,
  Srivatsan Srinivasan, Wojciech Stokowiec, and Fabio Viola.
\newblock The {D}eep{M}ind {JAX} {E}cosystem, 2020.
\newblock URL \url{http://github.com/deepmind}.

\bibitem[Bastani et~al.(2018)Bastani, Pu, and
  Solar-Lezama]{bastani2018verifiable}
Osbert Bastani, Yewen Pu, and Armando Solar-Lezama.
\newblock Verifiable reinforcement learning via policy extraction.
\newblock \emph{Advances in neural information processing systems}, 31, 2018.

\bibitem[{Bellemare} et~al.(2013){Bellemare}, {Naddaf}, {Veness}, and
  {Bowling}]{bellemare13arcade}
M.~G. {Bellemare}, Y.~{Naddaf}, J.~{Veness}, and M.~{Bowling}.
\newblock The arcade learning environment: An evaluation platform for general
  agents.
\newblock \emph{Journal of Artificial Intelligence Research}, 47:\penalty0
  253--279, jun 2013.

\bibitem[Bellemare et~al.(2013)Bellemare, Naddaf, Veness, and
  Bowling]{bellemare2013arcade}
Marc~G Bellemare, Yavar Naddaf, Joel Veness, and Michael Bowling.
\newblock The arcade learning environment: An evaluation platform for general
  agents.
\newblock \emph{Journal of Artificial Intelligence Research}, 47:\penalty0
  253--279, 2013.

\bibitem[Bellemare et~al.(2017)Bellemare, Dabney, and
  Munos]{bellemare2017distributional}
Marc~G Bellemare, Will Dabney, and R{\'e}mi Munos.
\newblock A distributional perspective on reinforcement learning.
\newblock In \emph{International Conference on Machine Learning}, pages
  449--458. PMLR, 2017.

\bibitem[Brown et~al.(2020)Brown, Mann, Ryder, Subbiah, Kaplan, Dhariwal,
  Neelakantan, Shyam, Sastry, Askell, et~al.]{brown2020language}
Tom Brown, Benjamin Mann, Nick Ryder, Melanie Subbiah, Jared~D Kaplan, Prafulla
  Dhariwal, Arvind Neelakantan, Pranav Shyam, Girish Sastry, Amanda Askell,
  et~al.
\newblock Language models are few-shot learners.
\newblock \emph{Advances in neural information processing systems},
  33:\penalty0 1877--1901, 2020.

\bibitem[Castro et~al.(2018)Castro, Moitra, Gelada, Kumar, and
  Bellemare]{castro2018dopamine}
Pablo~Samuel Castro, Subhodeep Moitra, Carles Gelada, Saurabh Kumar, and Marc~G
  Bellemare.
\newblock Dopamine: A research framework for deep reinforcement learning.
\newblock \emph{arXiv preprint arXiv:1812.06110}, 2018.

\bibitem[Chen et~al.(2021)Chen, Lu, Rajeswaran, Lee, Grover, Laskin, Abbeel,
  Srinivas, and Mordatch]{chen2021decision}
Lili Chen, Kevin Lu, Aravind Rajeswaran, Kimin Lee, Aditya Grover, Misha
  Laskin, Pieter Abbeel, Aravind Srinivas, and Igor Mordatch.
\newblock Decision transformer: Reinforcement learning via sequence modeling.
\newblock \emph{Advances in neural information processing systems}, 34, 2021.

\bibitem[Chowdhery et~al.(2022)Chowdhery, Narang, Devlin, Bosma, Mishra,
  Roberts, Barham, Chung, Sutton, Gehrmann, et~al.]{chowdhery2022palm}
Aakanksha Chowdhery, Sharan Narang, Jacob Devlin, Maarten Bosma, Gaurav Mishra,
  Adam Roberts, Paul Barham, Hyung~Won Chung, Charles Sutton, Sebastian
  Gehrmann, et~al.
\newblock Palm: Scaling language modeling with pathways.
\newblock \emph{arXiv preprint arXiv:2204.02311}, 2022.

\bibitem[Cuccu et~al.(2018)Cuccu, Togelius, and
  Cudr{\'e}-Mauroux]{cuccu2018playing}
Giuseppe Cuccu, Julian Togelius, and Philippe Cudr{\'e}-Mauroux.
\newblock Playing atari with six neurons.
\newblock \emph{arXiv preprint arXiv:1806.01363}, 2018.

\bibitem[Dabral et~al.(2021)Dabral, Ramakrishnan, Jyothi,
  et~al.]{dabral2021rudder}
Rishabh Dabral, Ganesh Ramakrishnan, Preethi Jyothi, et~al.
\newblock Rudder: A cross lingual video and text retrieval dataset.
\newblock \emph{arXiv preprint arXiv:2103.05457}, 2021.

\bibitem[Devlin et~al.(2018)Devlin, Chang, Lee, and Toutanova]{devlin2018bert}
Jacob Devlin, Ming-Wei Chang, Kenton Lee, and Kristina Toutanova.
\newblock Bert: Pre-training of deep bidirectional transformers for language
  understanding.
\newblock \emph{arXiv preprint arXiv:1810.04805}, 2018.

\bibitem[Dosovitskiy et~al.(2020)Dosovitskiy, Beyer, Kolesnikov, Weissenborn,
  Zhai, Unterthiner, Dehghani, Minderer, Heigold, Gelly,
  et~al.]{dosovitskiy2020image}
Alexey Dosovitskiy, Lucas Beyer, Alexander Kolesnikov, Dirk Weissenborn,
  Xiaohua Zhai, Thomas Unterthiner, Mostafa Dehghani, Matthias Minderer, Georg
  Heigold, Sylvain Gelly, et~al.
\newblock An image is worth 16x16 words: Transformers for image recognition at
  scale.
\newblock \emph{arXiv preprint arXiv:2010.11929}, 2020.

\bibitem[Espeholt et~al.(2018)Espeholt, Soyer, Munos, Simonyan, Mnih, Ward,
  Doron, Firoiu, Harley, Dunning, et~al.]{espeholt2018impala}
Lasse Espeholt, Hubert Soyer, Remi Munos, Karen Simonyan, Vlad Mnih, Tom Ward,
  Yotam Doron, Vlad Firoiu, Tim Harley, Iain Dunning, et~al.
\newblock Impala: Scalable distributed deep-rl with importance weighted
  actor-learner architectures.
\newblock In \emph{International Conference on Machine Learning}, pages
  1407--1416. PMLR, 2018.

\bibitem[Fujimoto et~al.(2019)Fujimoto, Meger, and Precup]{fujimoto2019off}
Scott Fujimoto, David Meger, and Doina Precup.
\newblock Off-policy deep reinforcement learning without exploration.
\newblock In \emph{International Conference on Machine Learning}, pages
  2052--2062. PMLR, 2019.

\bibitem[Furuta et~al.(2021{\natexlab{a}})Furuta, Kozuno, Matsushima, Matsuo,
  and Gu]{furuta2021co}
Hiroki Furuta, Tadashi Kozuno, Tatsuya Matsushima, Yutaka Matsuo, and
  Shixiang~Shane Gu.
\newblock Co-adaptation of algorithmic and implementational innovations in
  inference-based deep reinforcement learning.
\newblock \emph{Advances in neural information processing systems},
  34:\penalty0 9828--9842, 2021{\natexlab{a}}.

\bibitem[Furuta et~al.(2021{\natexlab{b}})Furuta, Matsuo, and
  Gu]{furuta2021generalized}
Hiroki Furuta, Yutaka Matsuo, and Shixiang~Shane Gu.
\newblock Generalized decision transformer for offline hindsight information
  matching.
\newblock \emph{arXiv preprint arXiv:2111.10364}, 2021{\natexlab{b}}.

\bibitem[Gelada et~al.(2019)Gelada, Kumar, Buckman, Nachum, and
  Bellemare]{gelada2019deepmdp}
Carles Gelada, Saurabh Kumar, Jacob Buckman, Ofir Nachum, and Marc~G Bellemare.
\newblock Deepmdp: Learning continuous latent space models for representation
  learning.
\newblock In \emph{International Conference on Machine Learning}, pages
  2170--2179. PMLR, 2019.

\bibitem[Gulcehre et~al.(2020)Gulcehre, Wang, Novikov, Paine, G{\'o}mez, Zolna,
  Agarwal, Merel, Mankowitz, Paduraru, et~al.]{gulcehre2020rl}
Caglar Gulcehre, Ziyu Wang, Alexander Novikov, Thomas Paine, Sergio G{\'o}mez,
  Konrad Zolna, Rishabh Agarwal, Josh~S Merel, Daniel~J Mankowitz, Cosmin
  Paduraru, et~al.
\newblock Rl unplugged: A suite of benchmarks for offline reinforcement
  learning.
\newblock \emph{Advances in Neural Information Processing Systems},
  33:\penalty0 7248--7259, 2020.

\bibitem[Gupta et~al.(2021)Gupta, Savarese, Ganguli, and
  Fei-Fei]{gupta2021embodied}
Agrim Gupta, Silvio Savarese, Surya Ganguli, and Li~Fei-Fei.
\newblock Embodied intelligence via learning and evolution.
\newblock \emph{Nature communications}, 12\penalty0 (1):\penalty0 1--12, 2021.

\bibitem[Hafner et~al.(2019{\natexlab{a}})Hafner, Lillicrap, Ba, and
  Norouzi]{hafner2019dream}
Danijar Hafner, Timothy Lillicrap, Jimmy Ba, and Mohammad Norouzi.
\newblock Dream to control: Learning behaviors by latent imagination.
\newblock In \emph{International Conference on Learning Representations},
  2019{\natexlab{a}}.

\bibitem[Hafner et~al.(2019{\natexlab{b}})Hafner, Lillicrap, Fischer, Villegas,
  Ha, Lee, and Davidson]{hafner2019learning}
Danijar Hafner, Timothy Lillicrap, Ian Fischer, Ruben Villegas, David Ha,
  Honglak Lee, and James Davidson.
\newblock Learning latent dynamics for planning from pixels.
\newblock In \emph{International conference on machine learning}, pages
  2555--2565. PMLR, 2019{\natexlab{b}}.

\bibitem[Hafner et~al.(2020)Hafner, Lillicrap, Norouzi, and
  Ba]{hafner2020mastering}
Danijar Hafner, Timothy~P Lillicrap, Mohammad Norouzi, and Jimmy Ba.
\newblock Mastering atari with discrete world models.
\newblock In \emph{International Conference on Learning Representations}, 2020.

\bibitem[Hessel et~al.(2018)Hessel, Modayil, Van~Hasselt, Schaul, Ostrovski,
  Dabney, Horgan, Piot, Azar, and Silver]{hessel2018rainbow}
Matteo Hessel, Joseph Modayil, Hado Van~Hasselt, Tom Schaul, Georg Ostrovski,
  Will Dabney, Dan Horgan, Bilal Piot, Mohammad Azar, and David Silver.
\newblock Rainbow: Combining improvements in deep reinforcement learning.
\newblock In \emph{Thirty-second AAAI conference on artificial intelligence},
  2018.

\bibitem[Hoffmann et~al.(2022)Hoffmann, Borgeaud, Mensch, Buchatskaya, Cai,
  Rutherford, Casas, Hendricks, Welbl, Clark, et~al.]{hoffmann2022training}
Jordan Hoffmann, Sebastian Borgeaud, Arthur Mensch, Elena Buchatskaya, Trevor
  Cai, Eliza Rutherford, Diego de~Las Casas, Lisa~Anne Hendricks, Johannes
  Welbl, Aidan Clark, et~al.
\newblock Training compute-optimal large language models.
\newblock \emph{arXiv preprint arXiv:2203.15556}, 2022.

\bibitem[Holtzman et~al.(2019)Holtzman, Buys, Du, Forbes, and
  Choi]{holtzman2019curious}
Ari Holtzman, Jan Buys, Li~Du, Maxwell Forbes, and Yejin Choi.
\newblock The curious case of neural text degeneration.
\newblock \emph{arXiv preprint arXiv:1904.09751}, 2019.

\bibitem[Huang et~al.(2020)Huang, Mordatch, and Pathak]{huang2020one}
Wenlong Huang, Igor Mordatch, and Deepak Pathak.
\newblock One policy to control them all: Shared modular policies for
  agent-agnostic control.
\newblock In \emph{International Conference on Machine Learning}, pages
  4455--4464. PMLR, 2020.

\bibitem[Jang et~al.(2022)Jang, Irpan, Khansari, Kappler, Ebert, Lynch, Levine,
  and Finn]{jang2022bc}
Eric Jang, Alex Irpan, Mohi Khansari, Daniel Kappler, Frederik Ebert, Corey
  Lynch, Sergey Levine, and Chelsea Finn.
\newblock Bc-z: Zero-shot task generalization with robotic imitation learning.
\newblock In \emph{Conference on Robot Learning}, pages 991--1002. PMLR, 2022.

\bibitem[Janner et~al.(2021)Janner, Li, and Levine]{janner2021offline}
Michael Janner, Qiyang Li, and Sergey Levine.
\newblock Offline reinforcement learning as one big sequence modeling problem.
\newblock \emph{Advances in neural information processing systems}, 34, 2021.

\bibitem[Kaiser et~al.(2017)Kaiser, Gomez, Shazeer, Vaswani, Parmar, Jones, and
  Uszkoreit]{kaiser2017one}
Lukasz Kaiser, Aidan~N Gomez, Noam Shazeer, Ashish Vaswani, Niki Parmar, Llion
  Jones, and Jakob Uszkoreit.
\newblock One model to learn them all.
\newblock \emph{arXiv preprint arXiv:1706.05137}, 2017.

\bibitem[Kalashnikov et~al.(2021)Kalashnikov, Varley, Chebotar, Swanson,
  Jonschkowski, Finn, Levine, and Hausman]{kalashnikov2021mt}
Dmitry Kalashnikov, Jacob Varley, Yevgen Chebotar, Benjamin Swanson, Rico
  Jonschkowski, Chelsea Finn, Sergey Levine, and Karol Hausman.
\newblock Mt-opt: Continuous multi-task robotic reinforcement learning at
  scale.
\newblock \emph{arXiv preprint arXiv:2104.08212}, 2021.

\bibitem[Kaplan et~al.(2020)Kaplan, McCandlish, Henighan, Brown, Chess, Child,
  Gray, Radford, Wu, and Amodei]{kaplan2020scaling}
Jared Kaplan, Sam McCandlish, Tom Henighan, Tom~B Brown, Benjamin Chess, Rewon
  Child, Scott Gray, Alec Radford, Jeffrey Wu, and Dario Amodei.
\newblock Scaling laws for neural language models.
\newblock \emph{arXiv preprint arXiv:2001.08361}, 2020.

\bibitem[Kappen et~al.(2012)Kappen, G{\'o}mez, and Opper]{kappen2012optimal}
Hilbert~J Kappen, Vicen{\c{c}} G{\'o}mez, and Manfred Opper.
\newblock Optimal control as a graphical model inference problem.
\newblock \emph{Machine learning}, 87\penalty0 (2):\penalty0 159--182, 2012.

\bibitem[Krause et~al.(2020)Krause, Gotmare, McCann, Keskar, Joty, Socher, and
  Rajani]{krause2020gedi}
Ben Krause, Akhilesh~Deepak Gotmare, Bryan McCann, Nitish~Shirish Keskar,
  Shafiq Joty, Richard Socher, and Nazneen~Fatema Rajani.
\newblock Gedi: Generative discriminator guided sequence generation.
\newblock \emph{arXiv preprint arXiv:2009.06367}, 2020.

\bibitem[Kumar et~al.(2019)Kumar, Peng, and Levine]{kumar2019reward}
Aviral Kumar, Xue~Bin Peng, and Sergey Levine.
\newblock Reward-conditioned policies.
\newblock \emph{arXiv preprint arXiv:1912.13465}, 2019.

\bibitem[Kumar et~al.(2020)Kumar, Zhou, Tucker, and
  Levine]{kumar2020conservative}
Aviral Kumar, Aurick Zhou, George Tucker, and Sergey Levine.
\newblock Conservative q-learning for offline reinforcement learning.
\newblock \emph{Advances in Neural Information Processing Systems},
  33:\penalty0 1179--1191, 2020.

\bibitem[Kurin et~al.(2020)Kurin, Igl, Rockt{\"a}schel, Boehmer, and
  Whiteson]{kurin2020my}
Vitaly Kurin, Maximilian Igl, Tim Rockt{\"a}schel, Wendelin Boehmer, and Shimon
  Whiteson.
\newblock My body is a cage: the role of morphology in graph-based incompatible
  control.
\newblock \emph{arXiv preprint arXiv:2010.01856}, 2020.

\bibitem[Lee et~al.(2020)Lee, Fischer, Liu, Guo, Lee, Canny, and
  Guadarrama]{lee2020predictive}
Kuang-Huei Lee, Ian Fischer, Anthony Liu, Yijie Guo, Honglak Lee, John Canny,
  and Sergio Guadarrama.
\newblock Predictive information accelerates learning in rl.
\newblock \emph{Advances in Neural Information Processing Systems},
  33:\penalty0 11890--11901, 2020.

\bibitem[Levine et~al.(2020)Levine, Kumar, Tucker, and Fu]{levine2020offline}
Sergey Levine, Aviral Kumar, George Tucker, and Justin Fu.
\newblock Offline reinforcement learning: Tutorial, review, and perspectives on
  open problems.
\newblock \emph{arXiv preprint arXiv:2005.01643}, 2020.

\bibitem[Lu et~al.(2021)Lu, Grover, Abbeel, and Mordatch]{lu2021pretrained}
Kevin Lu, Aditya Grover, Pieter Abbeel, and Igor Mordatch.
\newblock Pretrained transformers as universal computation engines.
\newblock \emph{arXiv preprint arXiv:2103.05247}, 2021.

\bibitem[Lyle et~al.(2021)Lyle, Rowland, Ostrovski, and Dabney]{lyle2021effect}
Clare Lyle, Mark Rowland, Georg Ostrovski, and Will Dabney.
\newblock On the effect of auxiliary tasks on representation dynamics.
\newblock In \emph{International Conference on Artificial Intelligence and
  Statistics}, pages 1--9. PMLR, 2021.

\bibitem[Lynch and Sermanet(2020)]{lynch2020language}
Corey Lynch and Pierre Sermanet.
\newblock Language conditioned imitation learning over unstructured data.
\newblock \emph{arXiv preprint arXiv:2005.07648}, 2020.

\bibitem[Mania et~al.(2018)Mania, Guy, and Recht]{mania2018simple}
Horia Mania, Aurelia Guy, and Benjamin Recht.
\newblock Simple random search provides a competitive approach to reinforcement
  learning.
\newblock \emph{arXiv preprint arXiv:1803.07055}, 2018.

\bibitem[McCarthy et~al.(2006)McCarthy, Minsky, Rochester, and
  Shannon]{mccarthy2006proposal}
John McCarthy, Marvin~L Minsky, Nathaniel Rochester, and Claude~E Shannon.
\newblock A proposal for the dartmouth summer research project on artificial
  intelligence, august 31, 1955.
\newblock \emph{AI magazine}, 27\penalty0 (4):\penalty0 12--12, 2006.

\bibitem[Mendonca et~al.(2021)Mendonca, Rybkin, Daniilidis, Hafner, and
  Pathak]{mendonca2021discovering}
Russell Mendonca, Oleh Rybkin, Kostas Daniilidis, Danijar Hafner, and Deepak
  Pathak.
\newblock Discovering and achieving goals via world models.
\newblock \emph{Advances in Neural Information Processing Systems}, 34, 2021.

\bibitem[Mnih et~al.(2013)Mnih, Kavukcuoglu, Silver, Graves, Antonoglou,
  Wierstra, and Riedmiller]{mnih2013playing}
Volodymyr Mnih, Koray Kavukcuoglu, David Silver, Alex Graves, Ioannis
  Antonoglou, Daan Wierstra, and Martin Riedmiller.
\newblock Playing atari with deep reinforcement learning.
\newblock \emph{arXiv preprint arXiv:1312.5602}, 2013.

\bibitem[Mnih et~al.(2015)Mnih, Kavukcuoglu, Silver, Rusu, Veness, Bellemare,
  Graves, Riedmiller, Fidjeland, Ostrovski, et~al.]{mnih2015human}
Volodymyr Mnih, Koray Kavukcuoglu, David Silver, Andrei~A Rusu, Joel Veness,
  Marc~G Bellemare, Alex Graves, Martin Riedmiller, Andreas~K Fidjeland, Georg
  Ostrovski, et~al.
\newblock Human-level control through deep reinforcement learning.
\newblock \emph{nature}, 518\penalty0 (7540):\penalty0 529--533, 2015.

\bibitem[Mnih et~al.(2016)Mnih, Badia, Mirza, Graves, Lillicrap, Harley,
  Silver, and Kavukcuoglu]{mnih2016asynchronous}
Volodymyr Mnih, Adria~Puigdomenech Badia, Mehdi Mirza, Alex Graves, Timothy
  Lillicrap, Tim Harley, David Silver, and Koray Kavukcuoglu.
\newblock Asynchronous methods for deep reinforcement learning.
\newblock In \emph{International conference on machine learning}, pages
  1928--1937. PMLR, 2016.

\bibitem[Nachum and Yang(2021)]{nachum2021provable}
Ofir Nachum and Mengjiao Yang.
\newblock Provable representation learning for imitation with contrastive
  fourier features.
\newblock \emph{Advances in Neural Information Processing Systems}, 34, 2021.

\bibitem[Oord et~al.(2018)Oord, Li, and Vinyals]{oord2018representation}
Aaron van~den Oord, Yazhe Li, and Oriol Vinyals.
\newblock Representation learning with contrastive predictive coding.
\newblock \emph{arXiv preprint arXiv:1807.03748}, 2018.

\bibitem[Ortega et~al.(2021)Ortega, Kunesch, Del{\'e}tang, Genewein, Grau-Moya,
  Veness, Buchli, Degrave, Piot, Perolat, et~al.]{ortega2021shaking}
Pedro~A Ortega, Markus Kunesch, Gr{\'e}goire Del{\'e}tang, Tim Genewein, Jordi
  Grau-Moya, Joel Veness, Jonas Buchli, Jonas Degrave, Bilal Piot, Julien
  Perolat, et~al.
\newblock Shaking the foundations: delusions in sequence models for interaction
  and control.
\newblock \emph{arXiv preprint arXiv:2110.10819}, 2021.

\bibitem[Ouyang et~al.(2022)Ouyang, Wu, Jiang, Almeida, Wainwright, Mishkin,
  Zhang, Agarwal, Slama, Ray, et~al.]{ouyang2022training}
Long Ouyang, Jeff Wu, Xu~Jiang, Diogo Almeida, Carroll~L Wainwright, Pamela
  Mishkin, Chong Zhang, Sandhini Agarwal, Katarina Slama, Alex Ray, et~al.
\newblock Training language models to follow instructions with human feedback.
\newblock \emph{arXiv preprint arXiv:2203.02155}, 2022.

\bibitem[Parisotto et~al.(2015)Parisotto, Ba, and
  Salakhutdinov]{parisotto2015actor}
Emilio Parisotto, Jimmy~Lei Ba, and Ruslan Salakhutdinov.
\newblock Actor-mimic: Deep multitask and transfer reinforcement learning.
\newblock \emph{arXiv preprint arXiv:1511.06342}, 2015.

\bibitem[Pomerleau(1991)]{pomerleau1991efficient}
Dean~A Pomerleau.
\newblock Efficient training of artificial neural networks for autonomous
  navigation.
\newblock \emph{Neural computation}, 3\penalty0 (1):\penalty0 88--97, 1991.

\bibitem[Radford et~al.(2021)Radford, Kim, Hallacy, Ramesh, Goh, Agarwal,
  Sastry, Askell, Mishkin, Clark, et~al.]{radford2021learning}
Alec Radford, Jong~Wook Kim, Chris Hallacy, Aditya Ramesh, Gabriel Goh,
  Sandhini Agarwal, Girish Sastry, Amanda Askell, Pamela Mishkin, Jack Clark,
  et~al.
\newblock Learning transferable visual models from natural language
  supervision.
\newblock In \emph{International Conference on Machine Learning}, pages
  8748--8763. PMLR, 2021.

\bibitem[Rae et~al.(2021)Rae, Borgeaud, Cai, Millican, Hoffmann, Song,
  Aslanides, Henderson, Ring, Young, et~al.]{rae2021scaling}
Jack~W Rae, Sebastian Borgeaud, Trevor Cai, Katie Millican, Jordan Hoffmann,
  Francis Song, John Aslanides, Sarah Henderson, Roman Ring, Susannah Young,
  et~al.
\newblock Scaling language models: Methods, analysis \& insights from training
  gopher.
\newblock \emph{arXiv preprint arXiv:2112.11446}, 2021.

\bibitem[Raffel et~al.(2019)Raffel, Shazeer, Roberts, Lee, Narang, Matena,
  Zhou, Li, and Liu]{raffel2019exploring}
Colin Raffel, Noam Shazeer, Adam Roberts, Katherine Lee, Sharan Narang, Michael
  Matena, Yanqi Zhou, Wei Li, and Peter~J Liu.
\newblock Exploring the limits of transfer learning with a unified text-to-text
  transformer.
\newblock \emph{arXiv preprint arXiv:1910.10683}, 2019.

\bibitem[Raghu et~al.(2021)Raghu, Unterthiner, Kornblith, Zhang, and
  Dosovitskiy]{raghu2021vision}
Maithra Raghu, Thomas Unterthiner, Simon Kornblith, Chiyuan Zhang, and Alexey
  Dosovitskiy.
\newblock Do vision transformers see like convolutional neural networks?
\newblock \emph{Advances in Neural Information Processing Systems},
  34:\penalty0 12116--12128, 2021.

\bibitem[Reed et~al.(2022)Reed, Zolna, Parisotto, Colmenarejo, Novikov,
  Barth-Maron, Gimenez, Sulsky, Kay, Springenberg, Eccles, Bruce, Razavi,
  Edwards, Heess, Chen, Hadsell, Vinyals, Bordbar, and
  de~Freitas]{gato2022deepmind}
Scott Reed, Konrad Zolna, Emilio Parisotto, Sergio~Gomez Colmenarejo, Alexander
  Novikov, Gabriel Barth-Maron, Mai Gimenez, Yury Sulsky, Jackie Kay,
  Jost~Tobias Springenberg, Tom Eccles, Jake Bruce, Ali Razavi, Ashley Edwards,
  Nicolas Heess, Yutian Chen, Raia Hadsell, Oriol Vinyals, Mahyar Bordbar, and
  Nando de~Freitas.
\newblock A generalist agent, 2022.
\newblock URL \url{https://arxiv.org/abs/2205.06175}.

\bibitem[Reid et~al.(2022)Reid, Yamada, and Gu]{reid2022can}
Machel Reid, Yutaro Yamada, and Shixiang~Shane Gu.
\newblock Can wikipedia help offline reinforcement learning?
\newblock \emph{arXiv preprint arXiv:2201.12122}, 2022.

\bibitem[Rusu et~al.(2015)Rusu, Colmenarejo, Gulcehre, Desjardins, Kirkpatrick,
  Pascanu, Mnih, Kavukcuoglu, and Hadsell]{rusu2015policy}
Andrei~A Rusu, Sergio~Gomez Colmenarejo, Caglar Gulcehre, Guillaume Desjardins,
  James Kirkpatrick, Razvan Pascanu, Volodymyr Mnih, Koray Kavukcuoglu, and
  Raia Hadsell.
\newblock Policy distillation.
\newblock \emph{arXiv preprint arXiv:1511.06295}, 2015.

\bibitem[Schmidhuber(2019)]{schmidhuber2019reinforcement}
Juergen Schmidhuber.
\newblock Reinforcement learning upside down: Don't predict rewards--just map
  them to actions.
\newblock \emph{arXiv preprint arXiv:1912.02875}, 2019.

\bibitem[Schrittwieser et~al.(2020)Schrittwieser, Antonoglou, Hubert, Simonyan,
  Sifre, Schmitt, Guez, Lockhart, Hassabis, Graepel,
  et~al.]{schrittwieser2020mastering}
Julian Schrittwieser, Ioannis Antonoglou, Thomas Hubert, Karen Simonyan,
  Laurent Sifre, Simon Schmitt, Arthur Guez, Edward Lockhart, Demis Hassabis,
  Thore Graepel, et~al.
\newblock Mastering atari, go, chess and shogi by planning with a learned
  model.
\newblock \emph{Nature}, 588\penalty0 (7839):\penalty0 604--609, 2020.

\bibitem[Shachter(1988)]{shachter1988probabilistic}
Ross~D Shachter.
\newblock Probabilistic inference and influence diagrams.
\newblock \emph{Operations research}, 36\penalty0 (4):\penalty0 589--604, 1988.

\bibitem[Srivastava et~al.(2019)Srivastava, Shyam, Mutz, Ja{\'s}kowski, and
  Schmidhuber]{srivastava2019training}
Rupesh~Kumar Srivastava, Pranav Shyam, Filipe Mutz, Wojciech Ja{\'s}kowski, and
  J{\"u}rgen Schmidhuber.
\newblock Training agents using upside-down reinforcement learning.
\newblock \emph{arXiv preprint arXiv:1912.02877}, 2019.

\bibitem[Su et~al.(2021)Su, Lu, Pan, Wen, and Liu]{su2021roformer}
Jianlin Su, Yu~Lu, Shengfeng Pan, Bo~Wen, and Yunfeng Liu.
\newblock Roformer: Enhanced transformer with rotary position embedding.
\newblock \emph{arXiv preprint arXiv:2104.09864}, 2021.

\bibitem[Todorov(2006)]{todorov2006linearly}
Emanuel Todorov.
\newblock Linearly-solvable markov decision problems.
\newblock \emph{Advances in neural information processing systems}, 19, 2006.

\bibitem[Toussaint(2009)]{toussaint2009robot}
Marc Toussaint.
\newblock Robot trajectory optimization using approximate inference.
\newblock In \emph{Proceedings of the 26th annual international conference on
  machine learning}, pages 1049--1056, 2009.

\bibitem[Tsimpoukelli et~al.(2021)Tsimpoukelli, Menick, Cabi, Eslami, Vinyals,
  and Hill]{tsimpoukelli2021multimodal}
Maria Tsimpoukelli, Jacob~L Menick, Serkan Cabi, SM~Eslami, Oriol Vinyals, and
  Felix Hill.
\newblock Multimodal few-shot learning with frozen language models.
\newblock \emph{Advances in Neural Information Processing Systems},
  34:\penalty0 200--212, 2021.

\bibitem[van~den Oord et~al.(2017)van~den Oord, Vinyals, and
  Kavukcuoglu]{van2017neural}
Aaron van~den Oord, Oriol Vinyals, and Koray Kavukcuoglu.
\newblock Neural discrete representation learning.
\newblock In \emph{Proceedings of the 31st International Conference on Neural
  Information Processing Systems}, pages 6309--6318, 2017.

\bibitem[Vaswani et~al.(2017)Vaswani, Shazeer, Parmar, Uszkoreit, Jones, Gomez,
  Kaiser, and Polosukhin]{vaswani2017attention}
Ashish Vaswani, Noam Shazeer, Niki Parmar, Jakob Uszkoreit, Llion Jones,
  Aidan~N Gomez, {\L}ukasz Kaiser, and Illia Polosukhin.
\newblock Attention is all you need.
\newblock \emph{Advances in neural information processing systems}, 30, 2017.

\bibitem[Xue et~al.(2021)Xue, Constant, Roberts, Kale, Al-Rfou, Siddhant,
  Barua, and Raffel]{xue2021mt5}
Linting Xue, Noah Constant, Adam Roberts, Mihir Kale, Rami Al-Rfou, Aditya
  Siddhant, Aditya Barua, and Colin Raffel.
\newblock mt5: A massively multilingual pre-trained text-to-text transformer.
\newblock In \emph{Proceedings of the 2021 Conference of the North American
  Chapter of the Association for Computational Linguistics: Human Language
  Technologies}, pages 483--498, 2021.

\bibitem[Yang and Klein(2021)]{yang2021fudge}
Kevin Yang and Dan Klein.
\newblock Fudge: Controlled text generation with future discriminators.
\newblock \emph{arXiv preprint arXiv:2104.05218}, 2021.

\bibitem[Yang and Nachum(2021)]{yang2021representation}
Mengjiao Yang and Ofir Nachum.
\newblock Representation matters: Offline pretraining for sequential decision
  making.
\newblock In \emph{International Conference on Machine Learning}, pages
  11784--11794. PMLR, 2021.

\bibitem[You et~al.(2019)You, Li, Reddi, Hseu, Kumar, Bhojanapalli, Song,
  Demmel, Keutzer, and Hsieh]{you2019large}
Yang You, Jing Li, Sashank Reddi, Jonathan Hseu, Sanjiv Kumar, Srinadh
  Bhojanapalli, Xiaodan Song, James Demmel, Kurt Keutzer, and Cho-Jui Hsieh.
\newblock Large batch optimization for deep learning: Training bert in 76
  minutes.
\newblock \emph{arXiv preprint arXiv:1904.00962}, 2019.

\bibitem[Yu et~al.(2020)Yu, Quillen, He, Julian, Hausman, Finn, and
  Levine]{yu2020meta}
Tianhe Yu, Deirdre Quillen, Zhanpeng He, Ryan Julian, Karol Hausman, Chelsea
  Finn, and Sergey Levine.
\newblock Meta-world: A benchmark and evaluation for multi-task and meta
  reinforcement learning.
\newblock In \emph{Conference on Robot Learning}, pages 1094--1100. PMLR, 2020.

\bibitem[Zhang et~al.(2020)Zhang, McAllister, Calandra, Gal, and
  Levine]{zhang2020learning}
Amy Zhang, Rowan McAllister, Roberto Calandra, Yarin Gal, and Sergey Levine.
\newblock Learning invariant representations for reinforcement learning without
  reconstruction.
\newblock \emph{arXiv preprint arXiv:2006.10742}, 2020.

\bibitem[Zheng et~al.(2022)Zheng, Zhang, and Grover]{zheng2022online}
Qinqing Zheng, Amy Zhang, and Aditya Grover.
\newblock Online decision transformer.
\newblock \emph{arXiv preprint arXiv:2202.05607}, 2022.

\end{thebibliography}
}
\section*{Checklist}

\begin{enumerate}

\item For all authors...
\begin{enumerate}
  \item Do the main claims made in the abstract and introduction accurately reflect the paper's contributions and scope?
    \answerYes{}
  \item Did you describe the limitations of your work?
    \answerYes{See Section~\ref{sec:conclusion}}
  \item Did you discuss any potential negative societal impacts of your work?
    \answerYes{See Section~\ref{sec:conclusion}}
  \item Have you read the ethics review guidelines and ensured that your paper conforms to them?
    \answerYes{}
\end{enumerate}

\item If you are including theoretical results...
\begin{enumerate}
  \item Did you state the full set of assumptions of all theoretical results?
    \answerNA{}
        \item Did you include complete proofs of all theoretical results?
    \answerNA{}
\end{enumerate}

\item If you ran experiments...
\begin{enumerate}
  \item Did you include the code, data, and instructions needed to reproduce the main experimental results (either in the supplemental material or as a URL)?
    \answerYes{}
  \item Did you specify all the training details (e.g., data splits, hyperparameters, how they were chosen)?
    \answerYes{}
        \item Did you report error bars (e.g., with respect to the random seed after running experiments multiple times)?
    \answerYes{}
        \item Did you include the total amount of compute and the type of resources used (e.g., type of GPUs, internal cluster, or cloud provider)?
    \answerYes{}
\end{enumerate}

\item If you are using existing assets (e.g., code, data, models) or curating/releasing new assets...
\begin{enumerate}
  \item If your work uses existing assets, did you cite the creators?
    \answerYes{}
  \item Did you mention the license of the assets?
    \answerNA{}
  \item Did you include any new assets either in the supplemental material or as a URL?
    \answerNA{}
  \item Did you discuss whether and how consent was obtained from people whose data you're using/curating?
    \answerNA{}
  \item Did you discuss whether the data you are using/curating contains personally identifiable information or offensive content?
    \answerNA{}
\end{enumerate}

\item If you used crowdsourcing or conducted research with human subjects...
\begin{enumerate}
  \item Did you include the full text of instructions given to participants and screenshots, if applicable?
    \answerNA{}
  \item Did you describe any potential participant risks, with links to Institutional Review Board (IRB) approvals, if applicable?
    \answerNA{}
  \item Did you include the estimated hourly wage paid to participants and the total amount spent on participant compensation?
    \answerNA{}
\end{enumerate}

\end{enumerate}

\newpage
\appendix
\section{Contribution Statement}

\textbf{Kuang-Huei Lee} Proposed project direction. Contributed to the JAX and TF Decision Transformer (DT) and BC code. Ran BC and DT experiments. Contributed to paper writing.

\textbf{Ofir Nachum} Proposed project direction. Contributed to TF code for DQN, CQL, and representation learning algorithms. Ran experiments for DQN and CQL. Contributed to paper writing.

\textbf{Mengjiao Yang} Contributed code in representation learning algorithms. Ran experiments for CPC, BERT, ACL pretraining + OOD finetuning. Contributed to paper writing.

\textbf{Lisa Lee} Contributed to TF codebase. Ran alternative environment experiments. Helped with paper editing.

\textbf{Daniel Freeman} Contributed to JAX training code and dataset generation pipelines. Generated datasets.  Helped with paper writing.

\textbf{Winnie Xu} Contributed to JAX training infrastructure, beginner / expert dataset generation, augmentation and metrics pipelines. Ran experiments for alternative DT variants. Helped with paper writing. Work done while a Research Intern.

\textbf{Sergio Guadarrama} Helped with project direction and experiment discussions. Worked on paper editing.

\textbf{Ian Fischer} Helped with project direction and experiment discussions. Worked on paper editing.

\textbf{Eric Jang} Helped with project direction, contributed to building JAX infrastructure. Helped with paper writing.

\textbf{Henryk Michalewski} Contributed to building jaxline training and data processing infrastructure. Ran fine-tuning experiments. Helped with project direction and experiment discussions. Contributed to paper writing.

\textbf{Igor Mordatch} Proposed project direction. Contributed to building JAX training and data processing infrastructure. Ran DT experiments. Contributed to paper writing and visualizations.

\section{Implementation Details}

\subsection{Transformer network architecture}
\label{subsec:architecture}

The input consists of a sequence of observations, returns, actions and rewards. Observations are images in the format $B \times T \times W \times H \times C$.
We use $84 \times 84$ grayscale images (\ie, $W=84, H=84, C=1$).
Similar to ViT~\citep{dosovitskiy2020image}, we extract $M$ non-overlapping image patches, perform a linear projection and then rasterise them into $d_{model}$-dimensional 1D tokens.
We define each patch to be $14 \times 14$ pixels (\ie, $M=6\times6=36$).
A learned positional embedding is added to each of the patch tokens $\obs_1, ..., \obs_M$ to retain positional information as in ViT.
As described in \cref{subsec:tokenization}, returns are discretized into 120 buckets in $\{-20, ..., 100\}$, and rewards are converted to ternary quantities $\{-1, 0, +1\}$.

For the whole sequence $\langle..., \obs^t_1, ..., \obs^t_M, \hat{\ret}^t, \act^t, \rew^t ,...\rangle$, we learn another positional embedding at each position and add to each token embedding. 
We experimented with rotary position embedding~\citep{su2021roformer}, but did not find a significant benefit from them in our setting.
On top of the token embeddings, our transformer models use a standard transformer decoder architecture.

A standard transformer implementation for sequence modeling would employ a sequential causal attention masking to prevent positions from attending to subsequent positions~\citep{vaswani2017attention}.
However, for the sequence $\langle..., \obs^t_1, ..., \obs^t_M, \hat{\ret}^t, \act^t, \rew^t ,...\rangle$ that we consider, we do not want to prevent the position corresponding to observation token $\obs^t_m$ from accessing subsequent observation tokens $\{\obs^t_{m'}: m' > m\}$ within the same timestep, since there is no clear sequential causal relation between image patches.
Therefore, we change the sequential causal masking to allow observation tokens within the same timestep to access each other, but not subsequent positions after $\obs^t_M$, \ie $\hat{\ret}^t, \act^t, \rew^t , \obs^{t+1}_1, ..., \obs^{t+1}_M, ...$

\Cref{tab:model_sizes} summarizes the transformer configurations we use for each model size.
We train these models on an internal cluster, each with 64 TPUv4.
Due to prohibitively long training times, we only evaluated one training seed.
For the 40M DT model, we perform four additional runs with different seeds to investigate the sensitivity to random seeds. 
The mean and standard deviation of the median score across 5 random seeds are 0.79 and 0.06; the mean and standard deviation of IQM scores are 0.95 and 0.05.
Compared to the differences to the baseline results and the results of other DT model sizes, we consider the variance to be relatively small.

\begin{table}[ht]
    \centering
    \begin{tabular}{lccccc}
        \hline
         Model & Layers & Hidden size $D$ ($d_{model})$ & Heads & Params & Training Time on 64 TPUv4 \\
        \hline
         DT-10M & 4 & 512 & 8 & 10M & 1 day \\
         DT-40M & 6 & 768 & 12 & 40M & 2 days \\
         DT-200M & 10 & 1280 & 20 & 200M & 8 days \\
         \hline
    \end{tabular}
    \caption{Multi-Game Decision Transformer Variants}
    \label{tab:model_sizes}
\end{table}

We performed hyperparameter search on DT-40M, and directly applied to DT-200M. 
With the set of hyperparameters working the best for smaller models, it still manifests a clear scaling trend of multi-game and transfer learning performance.

\subsection{Fine-tuning protocol for Atari games}
In the fine-tuning experiments, we reserved five games (Alien, MsPacman, Pong, Space Invaders and Star Gunner) to be used only for fine-tuning. 
These games were selected due to their varied gameplay characteristics. Each game was fine-tuned separately to measure the model's transfer performance for a fixed game. 
We use 1\% of the original dataset (corresponding to roughly $500\,000$ transitions) to specifically test fine-tuning in low-data regimes.

\subsection{Action and return sampling during in-game evaluation}
\label{subsec:appendix_eval}
We sample actions from the model with a temperature of 1. 
Inspired by Nucleus sampling (\citet{holtzman2019curious}), we only sample from the top 85th percentile action logits for all Decision Transformer models and Behavioral Cloning models (this parameter was selected to give highest performance for both models). 
While we train the model to predict actions for all timesteps in the sequence, during in-game evaluation, we execute the last predicted action in the sequence (conditioned on all past observations, and past generated actions, rewards, and target returns).

To generate target returns as discussed in \cref{subsec:expert_action_infer}, we sample them from the model with the temperature of 1 and the top 85th percentile logits. 
We use $\kappa = 10$ in all our experiments. 
To avoid storing the history of previously generated target returns (which may be difficult to incorporate into some RL frameworks), we experimented with autoregressively regenerating all target returns in the sequence, and found that to work well without requiring any special recurrent state maintenance outside of the model.
\Cref{alg:expert_inference} has pseudocode for our expert return and action inference.

\begin{algorithm}
\caption{\revision{Pseudocode for Expert Return and Action Inference (\Cref{subsec:expert_action_infer})}}\label{alg:expert_inference}
\begin{algorithmic}
\State \revision{Given an environment $E$, the current time step $t$. $\kappa=10$, the return upper bound $R_{high}=100$, the return lower bound $R_{low}=-20$}
\State \revision{Initialize a context window $\langle..., \rew^{t-2}, \obs^{t-1}_1, ..., \obs^{t-1}_M, \hat{\ret}^{t-1}, \act^{t-1}, \rew^{t-1}, \obs^{t}_1, ..., \obs^{t}_M\rangle$, abbreviated as $\dots$ in the following.}
\State \revision{\textit{\# Autoregressive return and action generation}}
\While{\revision{terminal state not reached yet}}
\State \revision{Compute 121 logits $(R^t=-20,\dots,100)$ for the categorical return distribution $P(R^t|...)$}
\State \revision{\textit{\# Increase logits proportionally to return magnitudes to prefer high magnitude}}
\State \revision{Define $\log P(R^t|\mathrm{expert}^t, ...) = \log P(R^t|...) + \kappa (R^t - R_{low})/(R_{high}-R_{low}) $}
\State \revision{\textit{\# Sample a return}}
\State \revision{$R^t \sim P(R^t|\mathrm{expert}^t, ...)$}
\State \revision{Compute logits for the categorical action distribution $P(a^t|R^t, ...)$}
\State \revision{\textit{\# Sample an action}}
\State \revision{$a^t \sim P(a^t|R^t, ...)$}
\State \revision{\textit{\# Interact with the environment}}
\State \revision{$\obs^{t+1}_1, ..., \obs^{t+1}_M, r^t \sim E_{step}(a^t)$}
\State \revision{$t = t + 1$ and shift the context window accordingly}
\EndWhile
\end{algorithmic}
\end{algorithm}

As an alternative way to generate expert-but-likely returns, we also experimented with simply generating $N$ return samples from the model according to log-probability $\log P_\theta(\ret^t|...)$, and picking the highest one. 
We then generate the action conditioned on this largest picked return as before. 
This avoids needing the hyperparameter $\kappa$. 
In this setting, we found $N=128$, inverse temperature of 0.75 for return sampling, no percentile cutoff for return sampling, and sampling from the top 50th percentile action logits with a temperature of 1 to work similarly well.

\subsection{Evaluation protocol and Atari environment details}
Our environment is the Atari 2600 Gym environment with pre-processing performed as in \citet{agarwal2020optimistic}. 
Our Atari observations are $84 \times 84$ grayscale images. 
We compress observation images to jpeg in the dataset (to keep dataset size small) and during in-game evaluation. 
All games use the same shared set of 18 discrete actions. 
For all methods, each game score is calculated by averaging over 16 model rollout episode trials. 
To reduce inter-trial variability, we do not use sticky actions during evaluation.

\subsection{Image augmentation}
\label{subsec:image_aug}
All models were trained with image augmentations. 
We investigate training with the following augmentation methods: random cropping, random channel permutation, random pixel permutation, horizontal flip, vertical flip, and random rotations. 
We found random cropping and random rotations to work the best. 
(In our random cropping implementation, images of size $84 \times 84$ are padded on each side with 4 zero-value pixels, and then randomly cropped to $84 \times 84$.)
In general, we aim to expand the domain of problems solved during training to similar kinds that we hope to generalize to by encoding useful inductive biases. 
We maintain the same random augmentation parameters for each window sequence. 
We apply data augmentation in both pre-training and fine-tuning.

\section{Baseline Implementation Details}

\paragraph{BC}
Our BC model is effectively the same as our DT model but removing the return token $\hat{R}^t$ from the training sequence:
\begin{equation*}
\label{eq:bc_sequence}
    x \; = \; \langle..., \obs^t_1, ..., \obs^t_M, \act^t, \rew^t ,...\rangle
\end{equation*}

Instead of predicting a return token (distribution) given observation tokens $\obs^t_1, ..., \obs^t_M$ and the previous part of the sequence, we directly predict an action token (distribution), which also means that we remove return conditioning for the BC model.
During evaluation, we sample actions with a temperature of 1, and sample from the top 85th percentile logits (as discussed in \cref{subsec:appendix_eval}).
All other implementation details and configurations are identical to DT.

\paragraph{C51 DQN} 
For single-game experiments, our implementation and training followed the details in~\citep{bellemare2017distributional} except for using multi-step learning with $n=4$. 
For multi-game experiments we trained using the details provided in the main text; we ran the algorithm for 15M gradient steps ($\approx 4$B environment steps $\approx16$B Atari frames).

\paragraph{CQL}
For CQL we use the same optimizer and learning rate as for C51 DQN. 
We use a per-replica batch size of 32 and run for 1M gradient steps on a TPU pod with 32 cores, yielding a global batch size of 256.
During finetuning for each game, we copy the entire $Q$-network trained with CQL, and apply an additional 100k gradient steps of batch size 32 on a single CPU, where each batch is sampled exclusively from the offline dataset of the finetuned game.
We also experimented with smaller learning rates ($0.00003$ instead of the default $0.00025$) and larger batch sizes ($1024$, $4096$) but found the results largely unchanged. 
We also tried using offline C51 and double DQN as opposed to CQL, and found performance to be worse.

\paragraph{CPC} 
For the CPC baseline~\citep{oord2018representation}, we apply a contrastive loss between $\phi(o_t),\phi(o_{t+1})$ using the objective function
\begin{equation}
    -\phi(o_{t+1})^\top W \phi(o_t) + \log \E_{\tilde s\sim\rho} [\exp\{\phi(\tilde{o})^\top W \phi(o_t)\}],
\end{equation}
where $W$ is a trainable matrix and $\rho$ is a non-trainable prior distribution; for mini-batch training we set $\rho$ to be the distribution of states in the mini-batch. 
The state representations $\phi(o)$ is parametrized by CNNs followed by two MLP layers with 512 units each interleaved with ReLU activation. 
For the CNN architecture, we used the C51 implementation with an Impala neural network architecture of three blocks using 16, 32, and 32 channels respectively, and trained with a batch size of 256 and learning rate of 0.00025 both during pretraining and downstream BC adaptation. 
We conduct representation learning for a total of 1M gradient steps, and finetune on 1\% data for 100k steps every 50k steps of representation learning and report the best finetuning results.

\paragraph{BERT and ACL} 
Our BERT and ACL baselines are based on the representation learning objectives described in~\citep{yang2021representation}.
For the BERT~\citep{devlin2018bert} state representation learning baseline, we 
(1) take a sub-trajectory $o_{t:t+k},a_{t:t+k},r_{t:t+k}$ from the dataset (without special tokenization as in DT), 
(2) randomly mask a subset of these, 
(3) pass the masked sequence into a transformer, and then 
(4) for each masked input state $o_{t+i}$, apply a contrastive loss %
between its representation $\phi(o_{t+i})$ and the transformer output $\mathrm{Transformer}[i]$ at the corresponding sequence position:
\begin{equation}
    -\phi(o_{t + i})^\top W\, \mathrm{Transformer}[i]  + \log \E_{\tilde o\sim\rho} [\exp\{ \phi(\tilde{o})^\top W\, \mathrm{Transformer}[i]\}],
\end{equation}
where $\rho$ is the distribution over states in the mini-batch.
For attentive contrastive learning (ACL)~\citep{yang2021representation}, we apply an additional action prediction loss to the output of BERT at the sequence positions of the action inputs.

To parameterize $\phi$, we use the same CNN architecture as in CPC, while the transformer is parameterized by two self-attention layers with 4 attention heads of 256 units each and feed-forward dimension 512. The transformer does not apply any additional directional masking to its inputs.
We used $k=16$.

Pretraining and finetuning is analogous to CPC. Namely, when finetuning we take the pretrained representation $\phi$ and use a BC objective for learning a neural network (two MLP layers with 512 units each) policy on top of this representation. %

\section{Comparisons between transformers and convolution networks}
\label{sec:udrl}

\begin{wrapfigure}{r}{0.45\linewidth}
    \vspace{-20pt}
     \centering
     \includegraphics[width=1\linewidth]{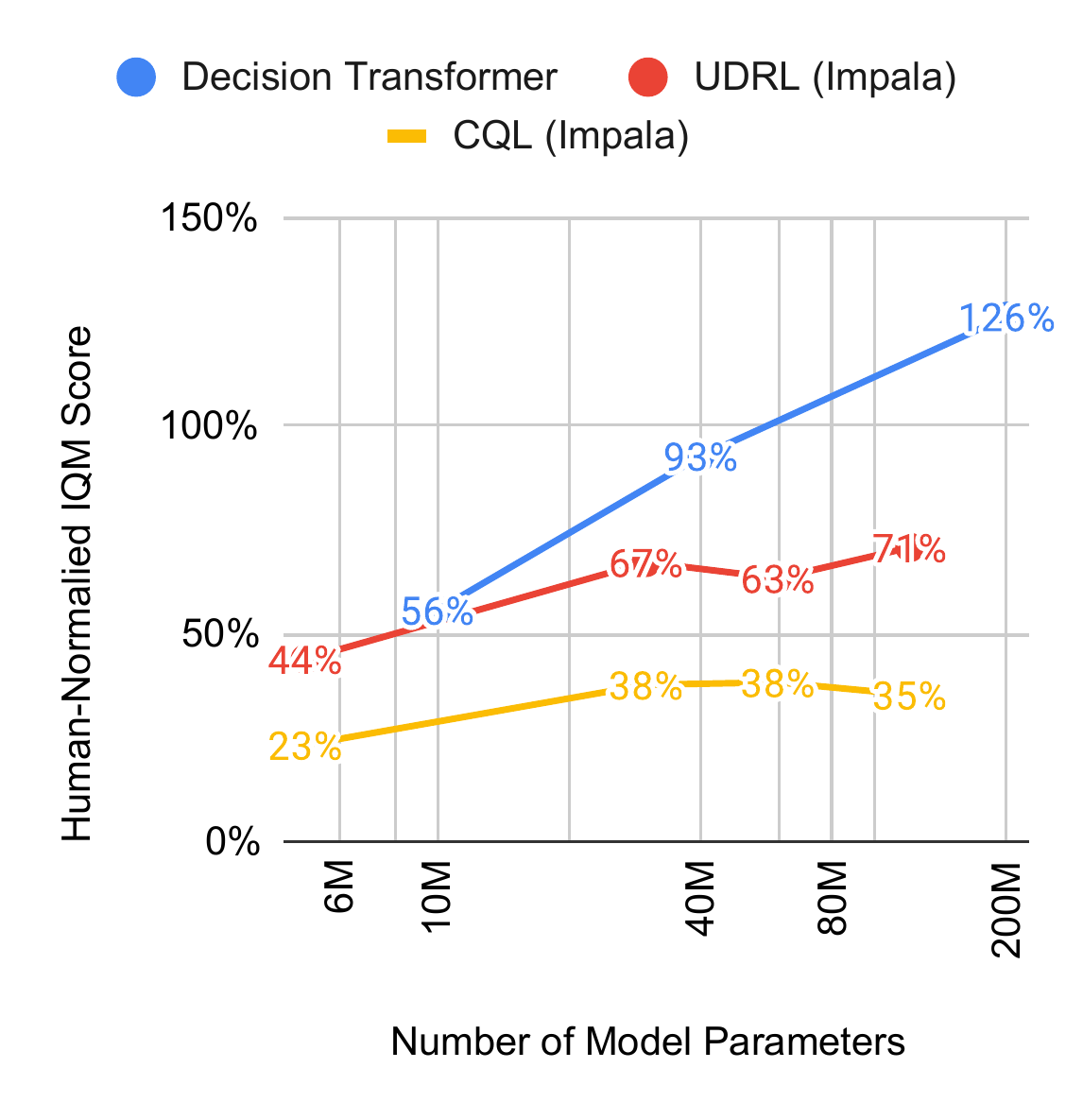}
     \caption{Performance scaling with model size for UDRL and CQL (Impala architecture) compared to Decision Transformer.}
     \label{fig:scaling_udrl}
\end{wrapfigure}

Decision Transformer is an Upside-Down RL (UDRL)~\citep{schmidhuber2019reinforcement,srivastava2019training} implementation that uses the transformer architecture and considers RL as a sequence modeling problem. 
To understand the benefit of the transformer architecture, we compare to a UDRL implementation that uses feed-forward, convolutional Impala networks~\citep{espeholt2018impala}.
Specifically, we use the same return, action, and reward tokenizers as in DT, and only replace the observation (four consecutive Atari frames stacked together) encoding to use the Impala architecture.
Similar to what we do for CQL, we also experiment with different sizes of the Impala architecture by varying the number of blocks and channels in each block of the Impala network: the number of blocks and channels is one of $(5~\mathrm{blocks}, 128~\mathrm{channels})\approx 5\mathrm{M~params}$, $(10~\mathrm{blocks}, 256~\mathrm{channels})\approx 30\mathrm{M~params}$, $(5~\mathrm{blocks}, 512~\mathrm{channels})\approx 60\mathrm{M~params}$.
We use a $(768, 768)$ $2$-layer fully-connected head to predict the next return token from observation embedding; another $(768, 768)$ head to predict the next action token from a concatenation of observation embedding and return token embedding; and another $(768, 768)$ head to predict the next reward token from a concatenation of observation embedding, return token embedding, and action token embedding.

The input to the model is slightly different from what we have for DT:
Instead of considering a $T$-timestep sub-trajectory ($T=4$) where each timestep contains $\obs^t, \ret^t, \act^t, \rew^t$, we stack $T$ image frames (as common in \cite{mnih2015human}), and only consider $\ret^t, \act^t, \rew^t$ from the last timestep.
All other design choices and evaluation protocols are the same as DT.

\cref{fig:scaling_udrl} shows clear advantages of Decision Transformer over UDRL with the Impala architecture.
In the comparison between UDRL (Impala) and CQL that uses the same Impala network at each model size we evaluated, we observe that UDRL (Impala) outperforms CQL.
The results show that the benefits of our method come not only from using network architectures, but also from the UDRL formulation.
\revision{While the reasons for the benefits of transformer architecture over other neural networks are still an open question in general (see e.g., \citep{raghu2021vision}), our hypothesis is that transformers allow for easier discovery of correlations between components of the input and output, due to the fact that transformers process the input as a flat sequence with attention allowed between any patch, action, or return token.}
Although it is not feasible to compare transformer with all possible convolutional architectures due to the broad design space, we believe these empirical results still show a clear trend favoring both UDRL and transformer architectures.

\section{Comparisons between methods using median human normalized scores}
\label{sec:median_scores}
We used inter-quartile mean (IQM) to aggregate performance over individual games in \cref{fig:method_scores}. 
Median is another metric commonly used to aggregate scores (although it has issues as discussed in \citep{agarwal2021deep}: it has high variability, and in the most extreme case, the median is unaffected by zero performance on nearly half of the tasks.). 
For completeness, we report the median scores for all methods:
\begin{figure}[ht]
    \centering
    \includegraphics[width=1.0\linewidth]{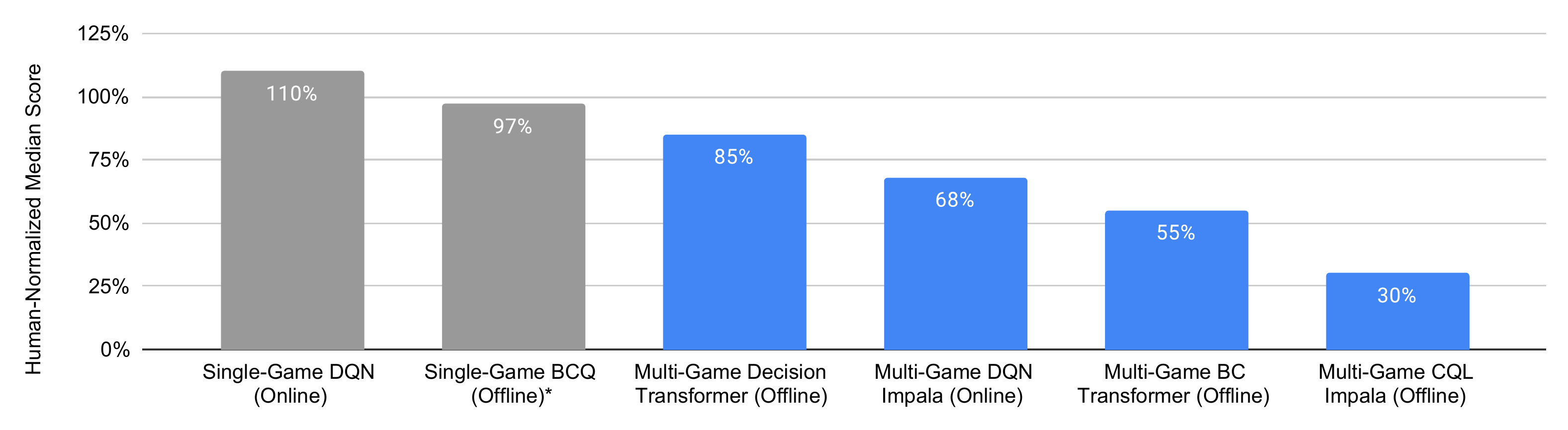}
    \caption{%
        Median human-normalized score across 41 Atari games. 
        Grey bars are single-game specialist models while blue are generalists. 
        Single-game BCQ results are from \citet{gulcehre2020rl}.
    }
    \label{fig:method_scores_median}
\end{figure}

\begin{figure}[ht]
     \centering
     \includegraphics[width=1\linewidth]{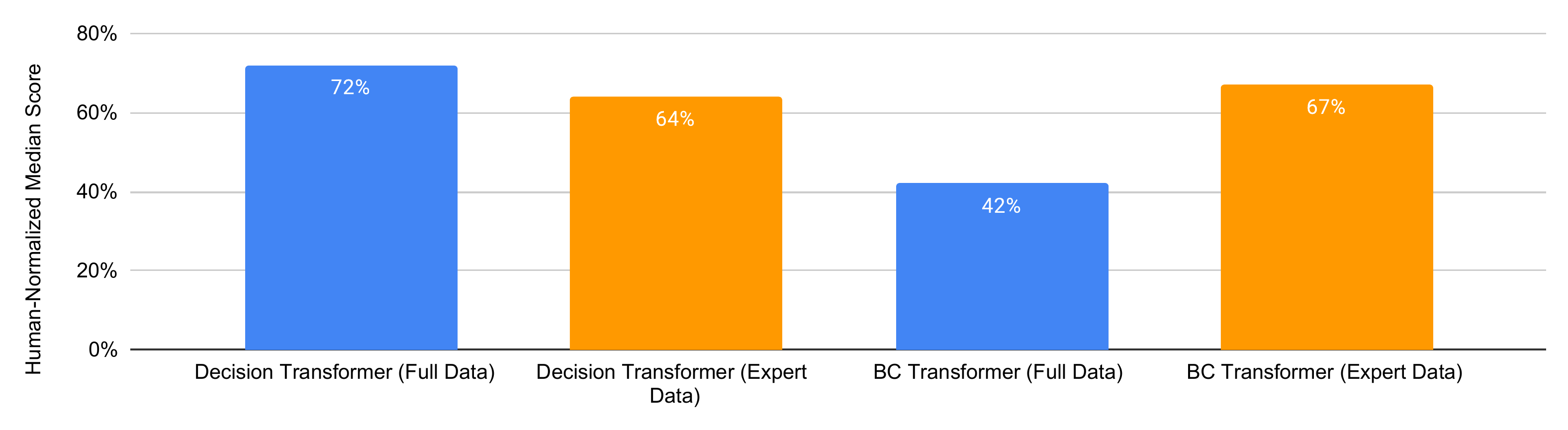}
     \caption{Median human-normalized scores of 40M transformer models trained on full data and only expert data.}
     \label{fig:expert_90p_comparison}
\end{figure}

For expert-filtering experiments in \cref{sec:exp_filtering}, we also provide the plot of expert filtering effects with median human-normalized scores in \cref{fig:expert_90p_comparison}. 
We note that ranking of various configurations do not change across aggregate metrics.

\begin{figure}[ht]
     \centering
     \includegraphics[width=0.5\linewidth]{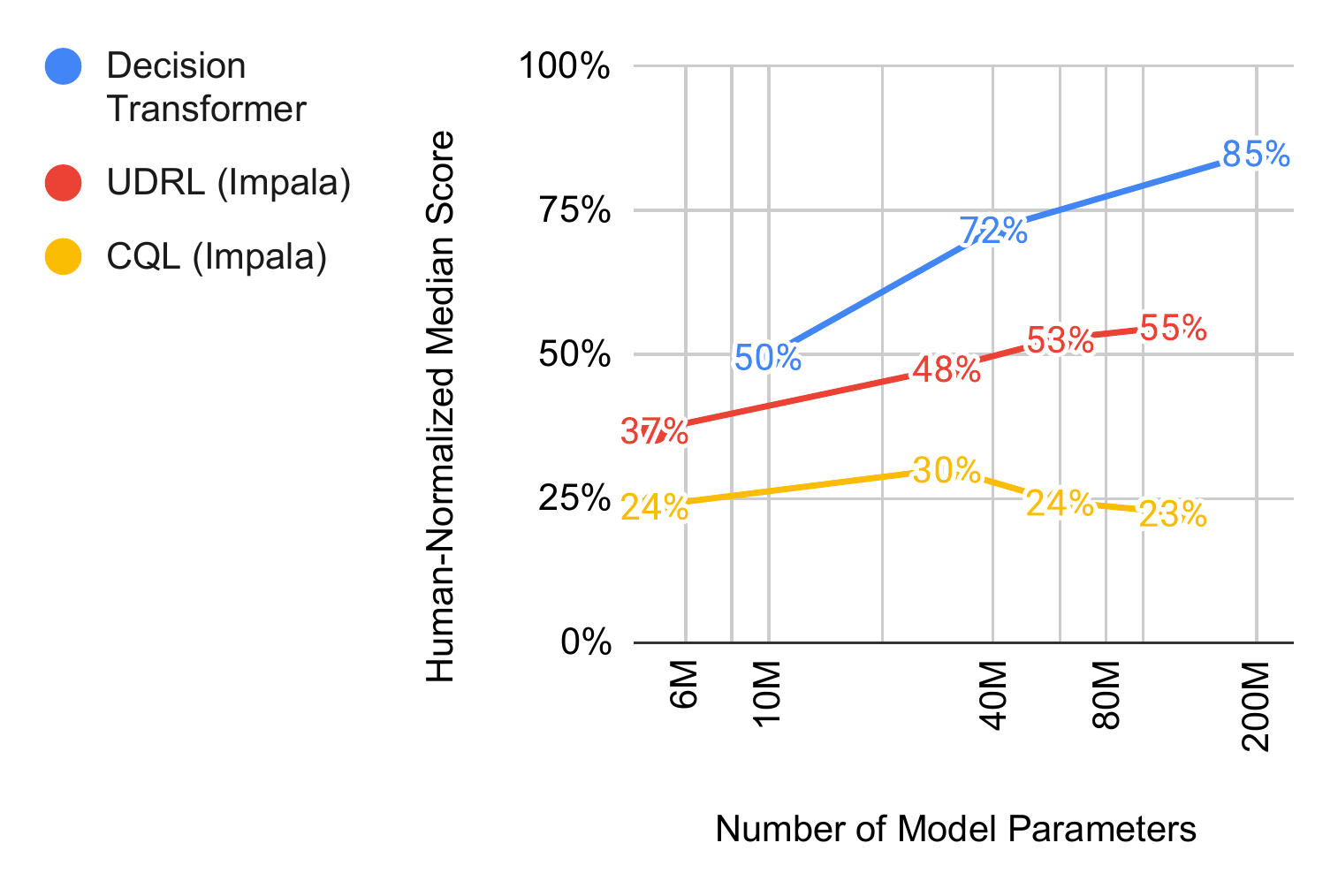}
     \caption{How UDRL (Impala architecture) median human-normalized score scales with model size on training set games, in comparisons with Decision Transformer and CQL (Impala architecture).}
     \label{fig:udrl_comparison_median}
\end{figure}

For Upside-Down RL comparison experiments \cref{sec:udrl}, we also provide median human-normalized scores in \cref{fig:udrl_comparison_median}.

\section{Details of Expert Dataset Generation}
\label{sec:filtering_appendix}
\begin{figure}[ht]
    \centering
     \begin{subfigure}{\textwidth}
         \centering
         \includegraphics[width=1\linewidth]{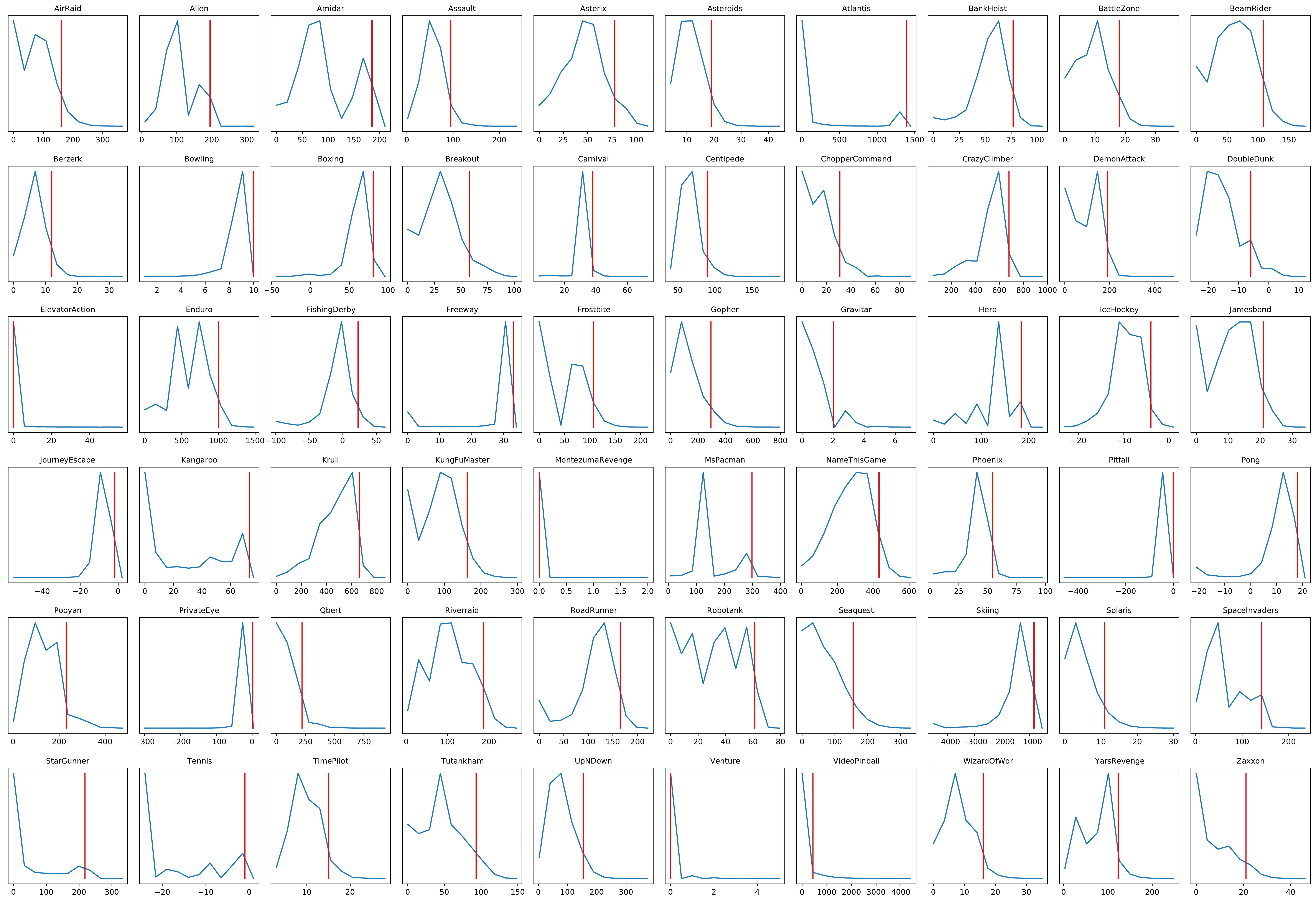}
     \end{subfigure}
     \caption{Histograms of rollout performance from \cite{agarwal2020optimistic} used to generate the expert dataset, with (unnormalized) score-density on the vertical axis, and game score (rewards are clipped) on the horizontal axis.  We indicate the 90th percentile performance cutoff with a red vertical line for each game.  Rollouts that exceeded this score threshold were included in the expert dataset.}
     \label{fig:atari_dataset}
\end{figure}

To generate the expert dataset for experiments in \cref{sec:exp_filtering}, we we filter our training data \cite{agarwal2020optimistic} from each game by episodic returns and only preserve top 10\% trajectories to produce an expert dataset. We plot of return histograms for reference in \cref{fig:atari_dataset}.

\section{Effect of Model Size on Training Speed}
\label{sec:model_size}

\begin{figure}[ht]
    \centering
     \begin{subfigure}{0.44\textwidth}
         \centering
         \includegraphics[width=1\linewidth]{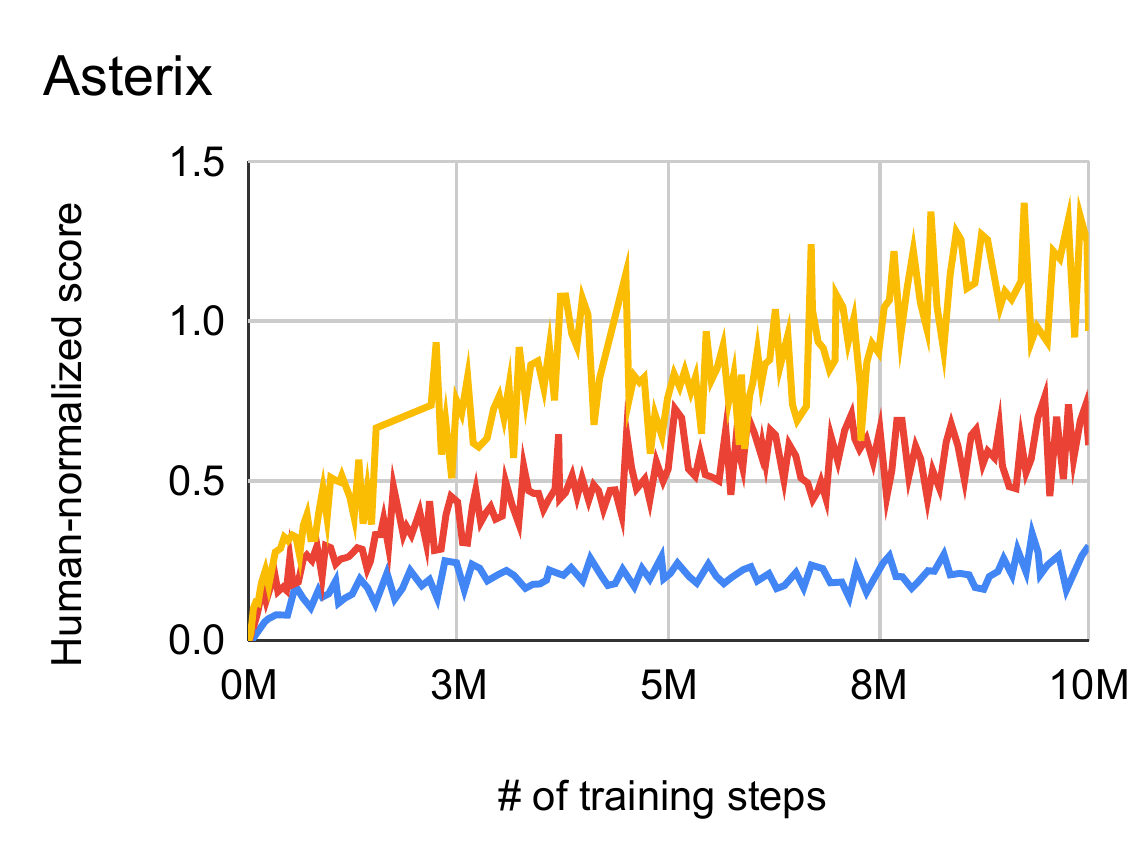}
     \end{subfigure}
     \hfill
     \begin{subfigure}{0.53\textwidth}
         \centering
         \includegraphics[width=1\linewidth]{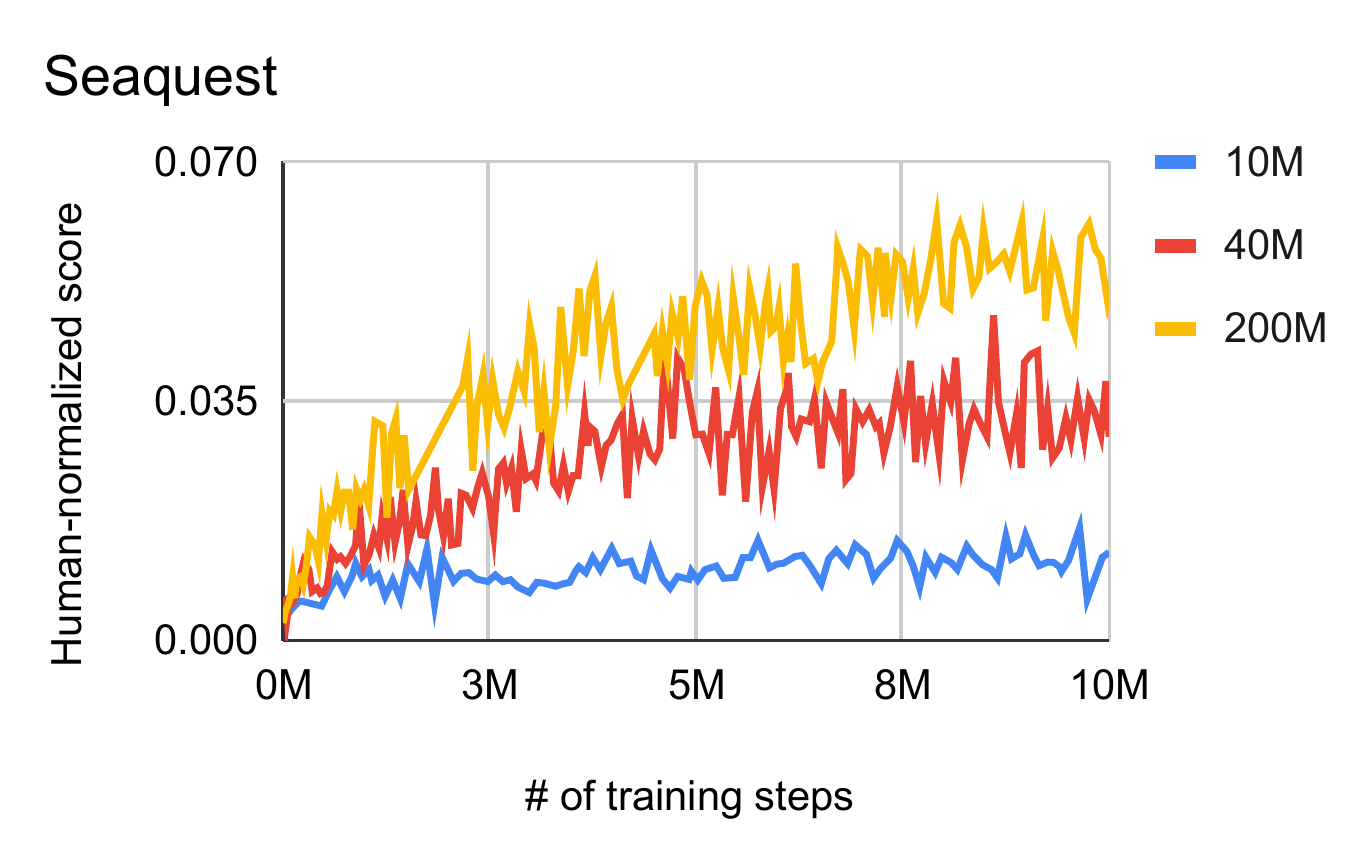}
     \end{subfigure}    
    \caption{Example game scores for different model sizes as multi-game training progresses.}
    \label{fig:progress}
\end{figure}

It is believed that large transformer-based language models train faster than smaller models, in the sense that they reach higher performance after observing a similar number of tokens~\citep{kaplan2020scaling,chowdhery2022palm}. 
We find this trend to hold in our setting as well. 
\cref{fig:progress} shows an example of performance on two example games as multi-game training progresses. 
We see that larger models reach higher scores per number of training steps taken (thus tokens observed). 

\section{Qualitative Attention Analysis}
\label{sec:attention}

\begin{figure}[ht]
    \centering
     \begin{subfigure}{0.3\textwidth}
         \centering
         \includegraphics[width=0.73\linewidth]{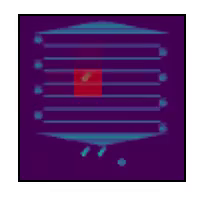}
         \caption{\emph{Asterix:} player}
     \end{subfigure}
     \hfill
     \begin{subfigure}{0.3\textwidth}
         \centering
         \includegraphics[width=0.73\linewidth]{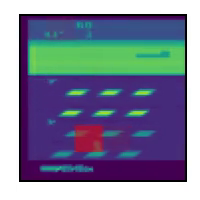}
         \caption{\emph{Frostbite:} player}
     \end{subfigure}
     \hfill
     \begin{subfigure}{0.3\textwidth}
         \centering
         \includegraphics[width=0.73\linewidth]{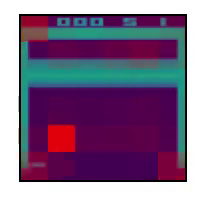}
         \caption{\emph{Breakout:} ball}
     \end{subfigure}     

     \begin{subfigure}{0.3\textwidth}
         \centering
         \includegraphics[width=0.73\linewidth]{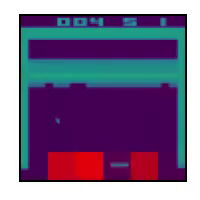}
         \caption{\emph{Breakout:} no paddle}
     \end{subfigure}
     \hfill
     \begin{subfigure}{0.3\textwidth}
         \centering
         \includegraphics[width=0.73\linewidth]{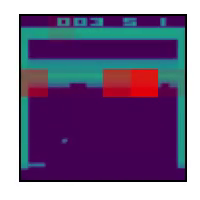}
         \caption{\emph{Breakout:} unbroken blocks}
     \end{subfigure}
     \hfill
     \begin{subfigure}{0.3\textwidth}
         \centering
         \includegraphics[width=0.73\linewidth]{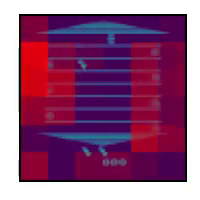}
         \caption{\emph{Asterix:} non-players}
     \end{subfigure}     
     \caption{Example image patches attended (red) for predicting next action by Decision Transformer.}     
     \label{fig:attention}
\end{figure}

We find that the Decision Transformer model consistently attends to observation image patches that contain meaningful game entities. 
\cref{fig:attention} visualizes selected attention heads and layers for various games.
We find heads consistently attend to entities such as player character, player's free movement space, non-player objects, and environment features.

\clearpage
\section{Raw Atari Scores}
We report full raw scores of 41 training Atari games for best performing sizes of multi-game models in \cref{tab:raw_scores}.

\begin{table}[h]
\begin{tabular}{l|rrrr}
\toprule
\bf{Game Name}      & \bf{DT (200M)}   & \bf{BC (200M)} & \bf{Online DQN (10M)} & \bf{CQL (60M)} \\
\hline
Amidar         & 101.5       & 101.0     & 629.8            & 4.0       \\
Assault        & 2,385.9     & 1,872.1   & 1,338.7          & 820.1     \\
Asterix        & 14,706.3    & 5,162.5   & 2,949.1          & 950.0     \\
Atlantis       & 3,105,342.3 & 4,237.5   & 976,030.4        & 16,800.0  \\
BankHeist      & 5.0         & 63.1      & 1,069.6          & 20.0      \\
BattleZone     & 17,687.5    & 9,250.0   & 26,235.2         & 5,000.0   \\
BeamRider      & 8,560.5     & 4,948.4   & 1,524.8          & 3,246.4   \\
Boxing         & 95.1        & 90.9      & 68.3             & 100.0     \\
Breakout       & 290.6       & 185.6     & 32.6             & 62.0      \\
Carnival       & 2,213.8     & 2,986.9   & 2,021.2          & 440.0     \\
Centipede      & 2,463.0     & 2,262.8   & 4,848.0          & 2,904.0   \\
ChopperCommand & 4,268.8     & 1,800.0   & 951.4            & 400.0     \\
CrazyClimber   & 126,018.8   & 123,350.0 & 146,362.5        & 139,300.0 \\
DemonAttack    & 23,768.4    & 7,870.6   & 446.8            & 1,202.0   \\
DoubleDunk     & -10.6       & -1.5      & -156.2           & -2.0      \\
Enduro         & 1,092.6     & 793.2     & 896.3            & 729.0     \\
FishingDerby   & 11.8        & 5.6       & -152.3           & 18.4      \\
Freeway        & 30.4        & 29.8      & 30.6             & 32.0      \\
Frostbite      & 2,435.6     & 782.5     & 2,748.4          & 408.0     \\
Gopher         & 9,935.0     & 3,496.3   & 3,205.6          & 700.0     \\
Gravitar       & 59.4        & 12.5      & 492.5            & 0.0       \\
Hero           & 20,408.8    & 13,850.0  & 26,568.8         & 14,040.0  \\
IceHockey      & -10.1       & -8.3      & -10.4            & -10.5     \\
Jamesbond      & 700.0       & 431.3     & 264.6            & 500.0     \\
Kangaroo       & 12,700.0    & 12,143.8  & 7,997.1          & 6,700.0   \\
Krull          & 8,685.6     & 8,058.8   & 8,221.4          & 7,170.0   \\
KungFuMaster   & 15,562.5    & 4,362.5   & 29,383.1         & 13,700.0  \\
NameThisGame   & 9,056.9     & 7,241.9   & 6,548.8          & 3,700.0   \\
Phoenix        & 5,295.6     & 4,326.9   & 3,932.5          & 1,880.0   \\
Pooyan         & 2,859.1     & 1,677.2   & 4,000.0          & 330.0     \\
Qbert          & 13,734.4    & 11,276.6  & 4,226.5          & 11,700.0  \\
Riverraid      & 14,755.6    & 9,816.3   & 7,306.6          & 3,810.0   \\
RoadRunner     & 54,568.8    & 49,118.8  & 25,233.0         & 50,900.0  \\
Robotank       & 63.2        & 44.6      & 9.2              & 17.0      \\
Seaquest       & 5,173.8     & 1,175.6   & 1,415.2          & 643.0     \\
TimePilot      & 2,743.8     & 1,312.5   & -883.1           & 2,400.0   \\
UpNDown        & 16,291.3    & 10,454.4  & 8,167.6          & 5,610.0   \\
VideoPinball   & 1,007.7     & 1,140.8   & 85,351.0         & 0.0       \\
WizardOfWor    & 187.5       & 443.8     & 975.9            & 500.0     \\
YarsRevenge    & 28,897.9    & 20,738.9  & 18,889.5         & 19,505.4  \\
Zaxxon         & 275.0       & 50.0      & -0.1             & 0.0 \\     
\bottomrule
\end{tabular}
\caption{Raw scores of 41 training Atari games for best performing multi-game models.}
\label{tab:raw_scores}
\end{table}

\end{document}